\documentclass[11pt,a4paper]{article}
\usepackage{tabu} 
\usepackage{bbm}
\usepackage{transparent}
\usepackage[hyperref]{acl2019} 

\usepackage{CJKutf8}

\aclfinalcopy

\usepackage{hyperref}
\usepackage{url} 

\usepackage{times}
\usepackage{latexsym}
\usepackage{xcolor}
\usepackage{balance}

\usepackage{url}

\usepackage{tikz}
\usepackage{tikz-qtree}
\usepackage{latexsym}
\usepackage{graphicx}			
\usepackage{times}
\usepackage{latexsym}
\usepackage{mathpartir}
\usepackage{pifont}

\usepackage{proof}
\usepackage{xspace}
\usepackage{multirow}
\usepackage{forest}
\usepackage{subcaption}

\usepackage{dsfont}
\usepackage{amsmath}
\usepackage{amssymb}
\usepackage{amsthm}
\usepackage{algorithm}
\newcommand{\namecite}[1]{\newcite{#1}}

\newcommand{\notes}[1]{}

\theoremstyle{definition}
\newtheorem{definition}{Definition}
\theoremstyle{plain}

\newcommand{\vecz}{\ensuremath{\mathbf{z}}}
\newcommand{\vech}{\ensuremath{\mathbf{h}}}

\newcommand{\ith}[1]{\ensuremath{i^{{th}}}}

\newcommand{\goesto}{\ensuremath{\rightarrow}\xspace}
\newcommand{\chn}[1]{\mbox{{\it {#1}}}}

\newcount\permx
\newcount\permy
\def\permdot#1#2{
\permx=#1 \advance\permx by-1
\permy=#2 \advance\permy by-1
\psframe[fillcolor=black, fillstyle=solid]
(\permx,\permy)(#1, #2)
}

\newcommand{\boxnum}[1]{{\setlength{\fboxsep}{1pt}\raisebox{1pt}{\hspace{1pt}\fbox{\tiny #1}\hspace{1pt}}}}
\newcommand{\ind}[1]{\ensuremath{_{\kern-0.5pt\boxnum{#1}}}}

\newcommand{\vecx}{\ensuremath{\mathbf{x}}\xspace}
\newcommand{\vecy}{\ensuremath{\mathbf{y}}\xspace}

\newcommand{\vecc}{\ensuremath{\mathbf{c}}\xspace}

\newcommand{\Bushi}{\chn{B\`ush\'i}\xspace}
\newcommand{\bushi}{\Bushi}

\newcommand{\yu}{\chn{y\v{u}}\xspace}

\newcommand{\huiwu}{\chn{hu\`iw\`u}\xspace}

\newcommand{\zongtong}{\chn{z\v{o}ngt\v{o}ng}\xspace}
\newcommand{\zai}{\chn{z\`ai}\xspace}
\newcommand{\mosike}{\chn{M\`os\-{i}k\={e}}\xspace}
\newcommand{\pujing}{\chn{P\v{u}j\={\i}ng}\xspace}

\newcommand{\meiguo}{\chn{M\v{e}igu\'o}\xspace}
\newcommand{\dangju}{\chn{d\=angj\'u}\xspace}
\newcommand{\dui}{\chn{du\`\i}\xspace}
\newcommand{\shate}{\chn{Sh\=at\`e}\xspace}
\newcommand{\jiezhe}{\chn{j\`{\i}zh\v{e}}\xspace}
\newcommand{\shizong}{\chn{sh\={\i}z\=ong}\xspace}
\newcommand{\danyou}{\chn{d\=any\=ou}\xspace}
\newcommand{\gandao}{\chn{g\v{a}nd\`ao}\xspace}
\newcommand{\an}{\chn{\`an}\xspace}
\newcommand{\yi}{\chn{y\=\i}\xspace}
\newcommand{\buman}{\chn{b\`um\v{a}n}\xspace}

\def\namecite{\newcite}

\newcommand{\smallnt}[1]{\ensuremath{_{\mbox{\tiny PP}}}\xspace}

\newcommand{\pseudocode}{Algorithm}
\floatname{algorithm}{\pseudocode}

\iffalse

\else

\fi

\newcommand{\eos}{\mbox{\scriptsize \texttt{<eos>}}\xspace}

\newcommand{\bajisitan}{B\=aj\={\i}s\={\i}t\v{a}n}
\newcommand{\yindu}{Y\`{\i}nd\`u}

\newcommand{\deen}{{{de}$\leftrightarrow${en}}\xspace}
\newcommand{\zhen}{{{zh}$\leftrightarrow${en}}\xspace}
\newcommand{\detoen}{{{de}$\goesto${en}}\xspace}
\newcommand{\entode}{{{en}$\goesto${de}}\xspace}
\newcommand{\zhtoen}{{{zh}$\goesto${en}}\xspace}
\newcommand{\entozh}{{{en}$\goesto${zh}}\xspace}

\newcommand{\floor}[1]{\lfloor #1 \rfloor}
\newcommand{\gwaitk}{\ensuremath{{g_\text{wait-$k$}}}\xspace}
\newcommand{\gcatchup}{\ensuremath{{g_\text{wait-$k$, $c$}}}\xspace}

\newcommand{\CW}{\ensuremath{\mathrm{CW}}\xspace}
\newcommand{\AP}{\ensuremath{\mathrm{AP}}\xspace}
\newcommand{\AL}{\ensuremath{\mathrm{AL}}\xspace}

\definecolor{chocolate}{rgb}{0.28, 0.02, 0.03}
\definecolor{PaleGreen}{rgb}{0.33, 0.545,0.33}
\definecolor{colorC0}{RGB}{51,113, 169}
\definecolor{colorC1}{RGB}{243,130,37}
\usepackage{array}
\usepackage{diagbox}

\title{STACL: Simultaneous Translation with Implicit Anticipation and \\ 
  Controllable Latency using Prefix-to-Prefix Framework
\thanks{\; 
  M.M.~and L.H.~contributed equally;
  L.H.~conceived the main ideas (prefix-to-prefix and wait-$k$)
  and directed the project,
  while M.M.~led the implementations on RNN and Transformer.
  See example videos, media reports,  code, and data at
 {\url{https://simultrans-demo.github.io/}}.
 } 
} 
\author{Mingbo Ma$^{1,3}$   \,
Liang Huang$^{1,3}$ \,
Hao Xiong$^2$ \,
Renjie Zheng$^{3}$\,
Kaibo Liu$^{1,3}$ \,
Baigong Zheng$^{1}$\,
\\
  {\bf
Chuanqiang Zhang$^2$ \,
Zhongjun He$^2$ \,
Hairong Liu$^1$ \,
Xing Li$^1$ \,
Hua Wu$^2$\,
Haifeng Wang$^2$
}
\\[0.1cm]
  $^{1}$Baidu Research, Sunnyvale, CA, USA
  \\ 
  $^2$Baidu, Inc., Beijing, China  
\\
  $^3$Oregon State University, Corvallis, OR, USA
\\
  {\tt \small \{mingboma, lianghuang, xionghao05, hezhongjun\}@baidu.com }
}

\begin{document}

\begin{CJK}{UTF8}{gbsn}
\maketitle
\begin{abstract}\vspace{-0.3cm}
  Simultaneous translation,
  which translates sentences before they are finished,
  is useful in many scenarios but is  notoriously difficult
  due to word-order differences. 
  While the conventional seq-to-seq framework is only suitable for full-sentence translation,
  we propose a novel prefix-to-prefix framework for simultaneous translation
  that implicitly learns to anticipate in a single translation model. 
  Within this framework, we present a very simple yet surprisingly effective ``wait-$k$'' policy
  {\em trained} to generate the target sentence concurrently with the source sentence, but always $k$ words behind.
  Experiments show our strategy achieves low latency and reasonable quality (compared to full-sentence translation) 
  on 4 directions: zh$\leftrightarrow$en and de$\leftrightarrow$en. 
\end{abstract}

\section{Introduction}

Simultaneous translation 
aims to automate simultaneous interpretation,
which translates concurrently with the source-language speech,
with a delay of only a few seconds.
This {\em additive} latency 
is much more desirable than the {\em multiplicative} 2$\times$ slowdown
in consecutive interpretation.


\begin{figure}
\centering
\includegraphics[width=7.8cm]{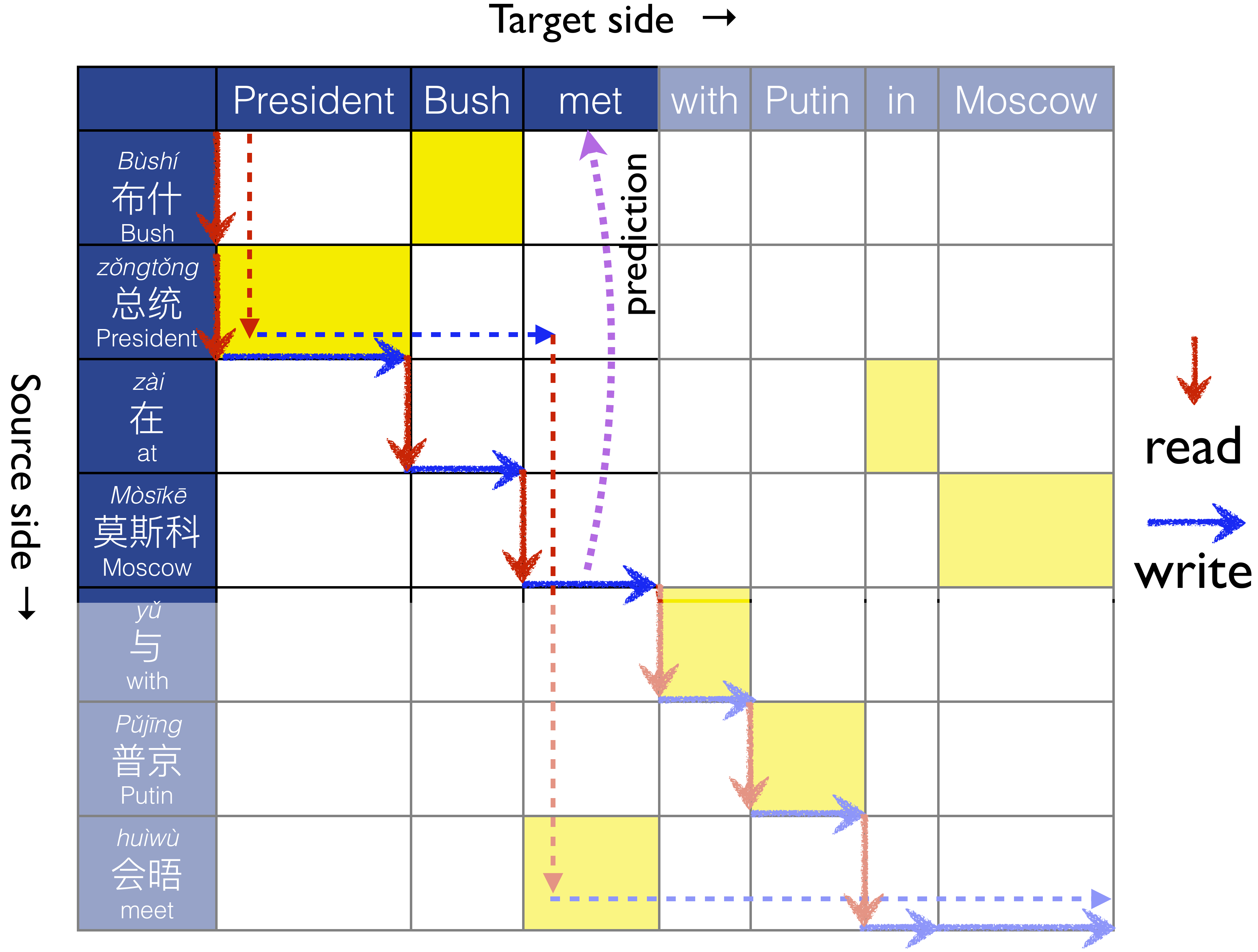}
\captionof{figure}{Our wait-$k$  model 
emits target word $y_t$ given source-side prefix $x_1 ...\, x_{t+k-1}$,
often before seeing the corresponding source word
(here $k$=2, outputing $y_3$=``met'' before $x_7$={\small ``\huiwu''}). 
Without anticipation, a 5-word wait is needed (dashed arrows). See also~Fig.~\ref{fig:idea2}.
}
\label{fig:idea}\vspace{-0.3cm}
\end{figure}
\begin{figure*}

\resizebox{\textwidth}{!}{
\setlength{\tabcolsep}{1.5pt}
\begin{tabu}{l | l l l l l l l l l l l l   l }
  \rowfont{\small}
& \textbf{\textit{B\`ush\'\i}}\, & \textbf{\textit{z\v{o}ngt\v{o}ng}} &\textbf{\textit{z\`ai}}  & \textbf\textit{M\`os\={\i}k\=e} & \transparent{0.5}\color{gray}\yu   &\transparent{0.5}\color{gray}\pujing &  \transparent{0.5}\color{blue}\huiwu \\
 &  \bf布什\, & \bf总统 &\bf在& \bf莫斯科 & \transparent{0.5}与 & \transparent{0.5}普京 & \transparent{0.5}\color{blue}会晤\\
 \rowfont{\small}
 & \bf Bush & \bf president& \bf in & \bf Moscow & \transparent{0.5}with/and & \transparent{0.5}Putin & \transparent{0.5}\color{blue}meet\\
\hline
(a) simultaneous: our wait-2  & \multicolumn{2}{c}{\small\em\color{red} ...wait 2 words...} &  pres.  & bush & \bf \color{blue}met &    \color{gray}with  &  
\color{gray}putin& \multicolumn{2}{l}{\color{gray}  in  moscow}\\
\hline
(b) non-simultaneous baseline & \multicolumn{7}{c}{\small\em\color{red} ..... wait whole sentence ......} & \multicolumn{2}{l}{pres. bush {\color{blue}met} with putin in moscow}\\
(c) simultaneous: test-time wait-2 & \multicolumn{2}{c}{\small\em\color{red} ...wait 2 words...} &  pres.  & bush & in & moscow &    and  &  \multicolumn{2}{l}{pol- ite meeting}\\
\hline
\hline
 &  布什\, & 总统 & &\qquad 在& 莫斯科 &与 &普京 & \color{blue}会晤\\
(d) simultaneous: non-predictive    & \multicolumn{2}{c}{\small\em\color{red} ...wait 2 words...} &  pres. bush\!\!\!\!\!\!\!\!\!\!\!\! & \multicolumn{5}{c}{{\em\color{red} \qquad ..... wait 5 words ......}} & {\color{blue}met}  with  putin   in  moscow\\
 \end{tabu}
}
\caption{Another view of  Fig.~\ref{fig:idea}, highlighting the prediction of  English {\color{blue}``met''}
  corresponding to the sentence-final Chinese verb {\color{blue}\huiwu}. 
(a) Our wait-$k$ policy (here $k=2$) translates concurrently with the source sentence, but always $k$ words behind. 
It correclty predicts the English verb given just the first 4 Chinese words (in bold),
lit.~``Bush president in Moscow'',
because it is trained in a prefix-to-prefix fashion (Sec.~\ref{sec:pred}), and the training data contains many prefix-pairs in the form of (X \zai Y ...,  X met ...).
(c) The test-time wait-$k$ decoding (Sec.~\ref{sec:waitk}) using the full-sentence model in (b) can not anticipate 
and produces nonsense translation.
(d) A simultaneous translator without anticipation such as \namecite{gu+:2017}
has to wait 5 words. 
\label{fig:idea2}
}
\end{figure*}

With this appealing property,
simultaneous interpretation
has been widely used in many scenarios including
multilateral organizations (UN/EU),
and international summits (APEC/G-20). 
However, due to the concurrent comprehension 
and production in two languages, 
it is  extremely challenging and exhausting  for humans:
the number of qualified
simultaneous interpreters worldwide is very limited,
and each can only last for about 15-30 minutes in one turn,
whose error rates grow exponentially after just minutes of interpreting \cite{moser+:1998}.
Moreover, 
limited memory forces
human interpreters to routinely omit source content \cite{hehe+:2016}.
Therefore, there is 
a critical need to 
develop simultaneous machine translation techniques
to reduce the burden of human interpreters 
and make it 
more accessible and affordable.

Unfortunately, simultaneous translation is also notoriously difficult for machines,
 due in large part to the diverging word order between the source and target languages.
For example, think about simultaneously translating an SOV 
language such as Japanese or German to an SVO language such as English or Chinese:\footnote{
Technically, German is SOV+V2 in main clauses,
and SOV in embedded clauses; Mandarin is a mix of SVO+SOV.}
you have to wait until the source language verb.
As a result, 
existing so-called ``real-time'' translation systems resort to conventional full-sentence translation,
causing an undesirable latency of at least one sentence. 
Some researchers, on the other hand,
have noticed the importance of verbs in SOV$\goesto$SVO translation
\cite{grissom+:2016}, and
have attempted to reduce latency by explicitly predicting the sentence-final German \cite{grissom+:2014} or English verbs \cite{matsubarayx+:2019},
which is limited to this particular case,
or unseen syntactic constituents \cite{oda+:2015,he+:2015},
which requires incremental parsing on the source sentence.
Some researchers propose to translate on an optimized sentence
segment level to get better translation 
accuracy \cite{oda+:2014,fujita+:2013,bangalore+:2012}.
More recently, \namecite{gu+:2017}
propose a two-stage model 
whose base model is a full-sentence model, 
On top of that, they use a READ/WRITE (R/W) model to decide, at every step, whether to wait for another source word (READ) or to emit a target word 
using the pretrained base model (WRITE),
and this R/W model is trained by reinforcement learning to prefer (rather than enforce) a specific latency, without updating the base model.
All these efforts have the following major limitations:
(a) none of them can achieve any arbitrary given latency such as ``3-word delay'';
(b) their base translation model is still trained on full sentences;
and (c) their systems are complicated, involving many components (such as pretrained model, prediction, and RL) and are difficult to train.


We instead present a very simple yet 
effective solution,
designing a novel prefix-to-prefix framework that predicts target words using only prefixes of the source sentence.
Within this framework, we study a special case, the ``wait-$k$'' policy, 
whose translation is always $k$ words behind the input.
Consider the Chinese-to-English example in Figs.~\ref{fig:idea}--\ref{fig:idea2}, where the translation of the sentence-final Chinese verb \huiwu (``meet'')
needs to be emitted earlier to avoid a long delay.
Our wait-2 model correctly anticipates the English verb given only the first 4 Chinese words (which provide enough clue for this prediction
given many similar prefixes in the training data).
We make the following contributions:
\begin{itemize}
    \setlength\itemsep{.1em}
\item Our prefix-to-prefix framework is tailored to simultaneous translation
and trained from scratch without using full-sentence models.
\item It
  seamlessly integrates implicit anticipation and translation in a single model
  that directly predicts {\em target} words without explictly hallucinating {\em source} ones.
\item As a special case, we present a  ``wait-$k$'' policy 
  that can satisfy any latency requirements.
\item This strategy 
  can be  applied to most sequence-to-sequence models with relatively minor changes.
  Due to space constraints, we only present its performance 
  the Transformer \cite{vaswani+:2017},
  though our initial experiments on RNNs \cite{bahdanau+:2014} showed equally strong results (see our November 2018 arXiv version {\footnotesize \url{https://arxiv.org/abs/1810.08398v3}}).
\item Experiments show our strategy achieves low latency and reasonable BLEU scores (compared to full-sentence translation baselines) 
  on 4 directions: \zhen and \deen. 
\end{itemize}

\section{Preliminaries: Full-Sentence NMT}
\label{sec:method}

We first briefly review 
standard (full-sentence) neural  translation to 
set up the notations. 

Regardless of the particular design of different seq-to-seq models,
the encoder always takes the input 
sequence $\vecx = (x_1,...,x_n)$ 
where each $x_i \in \mathbb{R}^{d_x}$ is a word embedding of $d_x$ dimensions,
and produces a new sequence of hidden states $\vech =f(\vecx) = (h_1,...,h_n)$. 
The encoding function $f$ can be implemented by RNN or Transformer. 

On the other hand, a (greedy) decoder 
predicts 
the next output word $y_t$
given the source sequence (actually its representation \vech) 
and previously generated words, denoted $\vecy_{<t}=(y_1,...,y_{t-1})$.
The decoder 
stops when it emits \eos,
and the final hypothesis $\vecy = (y_1,...,\eos)$ 
has probability 
\begin{equation}
p(\vecy \mid \vecx) = \textstyle\prod_{t=1}^{|\vecy|}  p(y_t \mid \vecx,\, \vecy_{<t})
\label{eq:gensentscore}
\end{equation}
At training time, we maximize the conditional probability of each ground-truth target sentence $\vecy^\star$ given input \vecx
over the whole training data $D$, or equivalently minimizing the following loss:
\begin{equation}
\ell(D) = - \textstyle\sum_{(\vecx,\vecy^\star)\in D} \log p(\vecy^\star \mid \vecx) 
\label{eq:train}
\end{equation}

\section{Prefix-to-Prefix and Wait-$k$ Policy} 
\label{sec:pred}

\begin{figure}
\centering
\includegraphics[width=7.8cm]{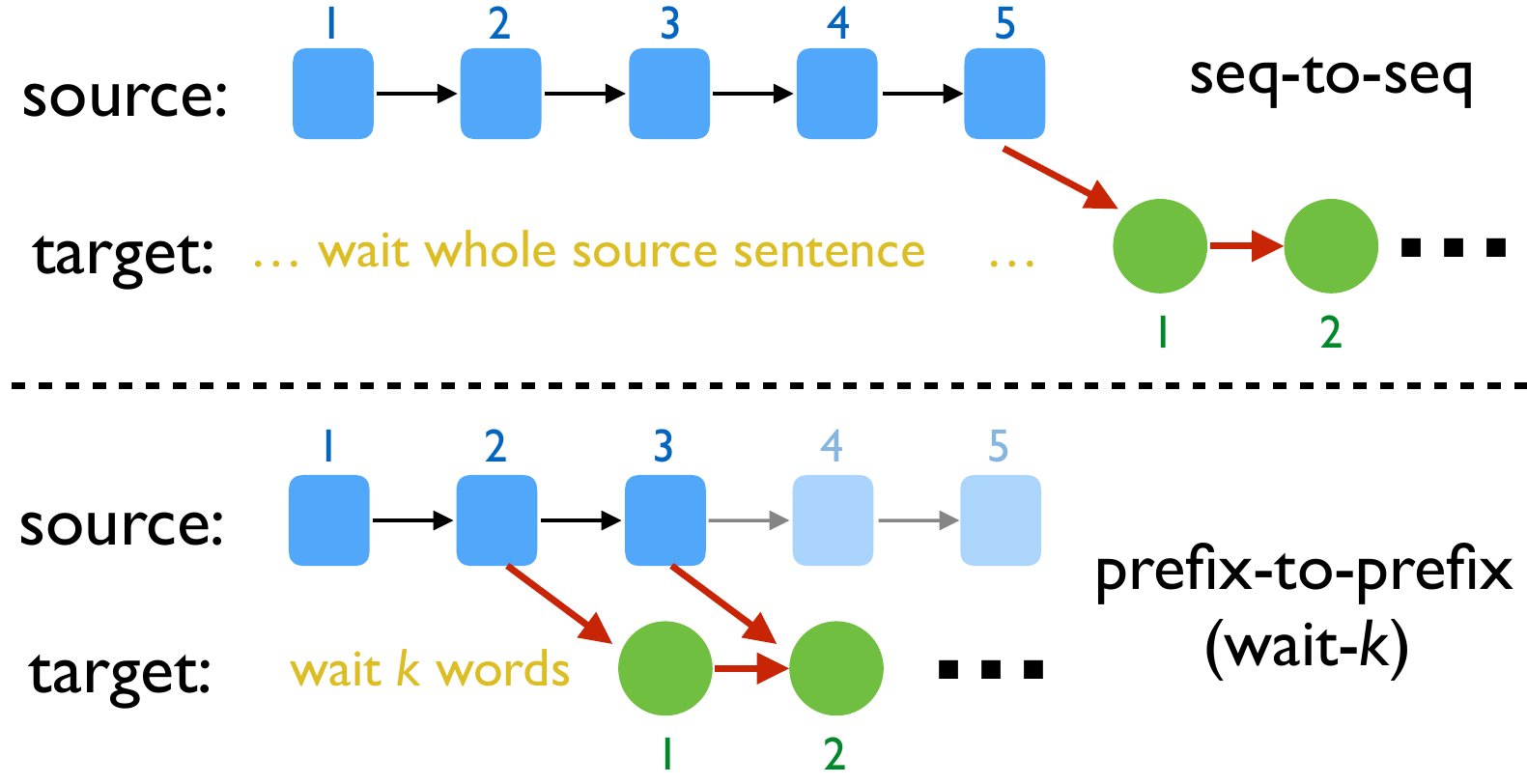}
\captionof{figure}{Seq-to-seq vs.~our prefix-to-prefix frameworks (showing wait-2 as an example).
}
\label{fig:p2p}
\end{figure}

In full-sentence translation (Sec.~\ref{sec:method}), 
each $y_i$ 
is predicted using the entire source sentence~$\vecx$.
But in simultaneous translation,
we need to translate concurrently with the (growing) source sentence,
so we design a new prefix-to-prefix architecture
to (be trained to) predict using a source prefix.

\subsection{Prefix-to-Prefix Architecture} 


\begin{definition}
Let $g(t)$ be a {\bf monotonic non-decreasing} function of $t$
that denotes the number of source words processed by the encoder when deciding the target word $y_t$.
\end{definition}
For example, in Figs.~\ref{fig:idea}--\ref{fig:idea2}, $g(3)=4$,
i.e., a 4-word Chinese prefix is used to predict $y_3$=``met''.
We use the source prefix 
$(x_1,...,x_{g(t)})$ rather than the whole $\vecx$
to predict $y_t$:
\(
p(y_t \mid \vecx_{\leq{g(t)}},\, \vecy_{<t}).
\)
Therefore the decoding probability is: 
\begin{equation}
\vspace{-0.1cm}
p_g(\vecy \mid \vecx) = \textstyle\prod_{t=1}^{|\vecy|} p(y_t \mid \vecx_{\leq{g(t)}},\, \vecy_{<t})
\label{eq:ourgoal}
\end{equation}
and given training $D$, the training objective is:
\begin{equation}
\ell_g(D) = - \textstyle\sum_{(\vecx,\,\vecy^\star)\in D} \log p_g(\vecy^\star \mid \vecx)
\label{eq:newtrain}
\end{equation}

Generally speaking, $g(t)$ can be used to represent any arbitrary policy,
and we give two special cases where $g(t)$ is constant:
(a) $g(t)=|\vecx|$: baseline full-sentence translation;
(b) $g(t)=0$: an ``oracle'' that does not rely on any source information.
Note that in any case, $0 \leq g(t)\leq |\vecx|$ for all $t$.

\iftrue
\begin{definition}
We define the {\bf ``cut-off'' step}, $\tau_g(|\vecx|)$,
to be the decoding step when source sentence finishes:
\begin{equation}
\tau_g(|\vecx|) = \min \{t \mid g(t)=|\vecx|\} 
\end{equation}
\end{definition}
\noindent
For example, in Figs.~\ref{fig:idea}--\ref{fig:idea2}, the cut-off step is 6,
i.e., the Chinese sentence finishes right before $y_6$=``in''.
\fi

\smallskip

\noindent
{\bf Training vs.~Test-Time Prefix-to-Prefix.}
While most previous work in simultaneous translation, in particular \namecite{bangalore+:2012} and \namecite{gu+:2017}, might be seen as special cases in this framework,
we note that only their {\em decoders} are prefix-to-prefix, while their training is still full-sentence-based. 
In other words, they use a full-sentence translation model to do simultaneous decoding, which is a mismatch between training and testing.
The essence of our idea, however, is to {\em train} the model to predict using source prefixes. 
Most importantly, 
this new training implicitly learns anticipation as a by-product, overcoming word-order differences such as SOV$\goesto$SVO.
Using the example in Figs.~\ref{fig:idea}--\ref{fig:idea2},
the anticipation of the English verb is possible because
the training data contains many prefix-pairs in the form of (X \zai Y ...,  X met ...),
thus although the prefix $\vecx_{\le 4}$=``\bushi \zongtong \zai \mosike'' (lit.~``Bush president in Moscow'') does not contain the verb,
it still provides enough clue to predict ``met''. 

\subsection{Wait-$k$ Policy}
\label{sec:waitk}

As a very simple example within the prefix-to-prefix framework,
we present a wait-$k$ policy,
which first wait $k$ source words,
and then translates concurrently with the rest of source sentence, i.e., the output is always $k$ words behind the input. 
This is inspired by human simultaneous interpreters who generally start translating a few seconds into the speakers' speech,
and finishes a few seconds after the speaker finishes.
For example, if $k=2$, the first target word is predicted using the first 2 source words,
and the second target word using the first 3 source words, etc; see Fig.~\ref{fig:p2p}.
More formally, its $g(t)$ is defined as follows:
\begin{equation}
  \gwaitk(t) =\min\{k+t-1, \, |\vecx|\}
\label{eq:policy}
\end{equation}
For this policy, the cut-off point $\tau_\gwaitk(|\vecx|)$ is exactly $|\vecx|-k+1$
(see Fig.~\ref{fig:catchup}).
From this step on,
$\gwaitk(t)$ is fixed to $|\vecx|$, 
which means the remaining target words (including this step) are generated using the full source sentence,
similar to conventional MT.
We call this part of output, $\vecy_{\geq |\vecx|-k}$, the ``tail'', and
can perform beam search on it (which we call ``tail beam search''),
but all earlier  words 
are generated greedily one by one 
(see Appendix). 

\smallskip
\noindent
    {\bf Test-Time Wait-$k$.}
    As an example of test-time prefix-to-prefix in the above subsection,
    we present a very simple ``test-time wait-$k$'' method, i.e.,
    using a full-sentence model but decoding it with a wait-$k$ policy (see also Fig.~\ref{fig:idea2}(c)).
    Our experiments show that this method, without the anticipation capability, performs much worse than our genuine wait-$k$
    when $k$ is small, but gradually catches up, and eventually both methods approach the full-sentence baseline ($k=\infty$).

\section{New Latency Metric: Average Lagging}
\label{sec:al}
Beside translation quality, latency is another crucial aspect for evaluating simultaneous translation. 
We first review existing latency metrics, highlighting their limitations,
aand then propose our new latency metric that address these limitations.

\subsection{Existing Metrics: CW and AP}

Consecutive Wait (CW) \cite{gu+:2017} is the number of source words
waited between two target words. Using our notation,
for a policy $g(\cdot)$, the per-step CW at step $t$ is
\(
\CW_g(t) = g(t) - g(t-1).
\)
The CW of a sentence-pair $(\vecx, \vecy)$ is the average CW over all consecutive wait segments:
\[
\resizebox{.48\textwidth}{!}{
  $\CW_g(\vecx,\vecy) = \displaystyle\frac{\sum_{t=1}^{|\vecy|} \CW_g(t)}{\sum_{t=1}^{|\vecy|} \mathbbm{1}_{\CW_g(t) > 0}} = \frac{|\vecx|}{\sum_{t=1}^{|\vecy|} \mathbbm{1}_{\CW_g(t) > 0}}$
  }
\]
\vspace{-0.3cm}

In other words, CW measures the average source segment length (the best case is 1 for word-by-word translation or our wait-1 and the worst case is $|\vecx|$ for full-sentence MT).
The drawback of CW is 
that CW is local latency measurement which is 
insensitive to the actual lagging behind.

Another latency measurement, Average Proportion (AP) \cite{Cho+:16} measures
the proportion of the area above a policy path in Fig.~\ref{fig:idea}:
\vspace{-0.2cm}
\begin{equation}
\AP_g(\vecx,\vecy) = \frac{1}{|\vecx|\;|\vecy|} \textstyle\sum_{t=1}^{|\vecy|} g(t) 
\label{eq:AP}
\end{equation}

AP has two major flaws:
First, it is sensitive to input length. 
For example, consider our wait-1 policy.
When $|\vecx|=|\vecy|=1$, AP is 1,
and when $|\vecx|=|\vecy|=2$, AP is 0.75,
and eventually AP approaches 0.5 when  $|\vecx|=|\vecy|\goesto\infty$.
However, in all these cases, there is a one word delay,
so AP is not fair between long and short sentences.
Second, being a percentage,
it is not obvious to the user the actual delays in number of words.

\if
Consecutive Wait (CW)  \cite{gu+:2017} measures the average lengths of conseuctive wait segments
(the best case is 1 for our wait-1 and the worst case is $|\vecx|$ for full-sentence MT).
For a policy $g(\cdot)$, the per-step CW at step $t$ is
\(
\CW_g(t) = g(t) - g(t-1).
\)
The CW of a sentence-pair $(\vecx, \vecy)$ is the average CW over all consecutive wait segments:
\[
\resizebox{.48\textwidth}{!}{
  $\CW_g(\vecx,\vecy) = \displaystyle\frac{\sum_{t=1}^{|\vecy|} \CW_g(t)}{\sum_{t=1}^{|\vecy|} \mathbbm{1}_{\CW_g(t) > 0}} = \frac{|\vecx|}{\sum_{t=1}^{|\vecy|} \mathbbm{1}_{\CW_g(t) > 0}}$
  }
\]
\vspace{-0.3cm}

Another latency measurement, Average Proportion (AP) \cite{Cho+:16} measures
the proportion of the area above a policy path in Fig.~\ref{fig:idea}:
\vspace{-0.2cm}
\begin{equation}
\AP_g(\vecx,\vecy) = \frac{1}{|\vecx|\;|\vecy|} \textstyle\sum_{t=1}^{|\vecy|} g(t) 
\label{eq:AP}
\end{equation}

The above

Another latency measurement, Average Proportion (AP) \cite{Cho+:16} measures
the proportion of the shaded area for a policy in Fig.~\ref{fig:catchup}:
\begin{equation}
\AP_g(\vecx,\vecy) = \frac{1}{|\vecx|\;|\vecy|} \textstyle\sum_{t=1}^{|\vecy|} g(t) 
\label{eq:AP}
\end{equation}



AP has two major flaws:
First, it is sensitive to input length. 
For example, consider our wait-1 policy.
When $|\vecx|=|\vecy|=1$, AP is 1,
and when $|\vecx|=|\vecy|=2$, AP is 0.75,
and eventually AP approaches 0.5 when  $|\vecx|=|\vecy|\goesto\infty$.
However, in all these cases, there is a one word delay,
so AP is not fair between long and short sentences.
Second, being a percentage,
it is not obvious to the user the actual delays in number of words.

\fi 

\subsection{New Metric: Average Lagging}

\begin{figure}
\centering
\includegraphics[width=7.8cm]{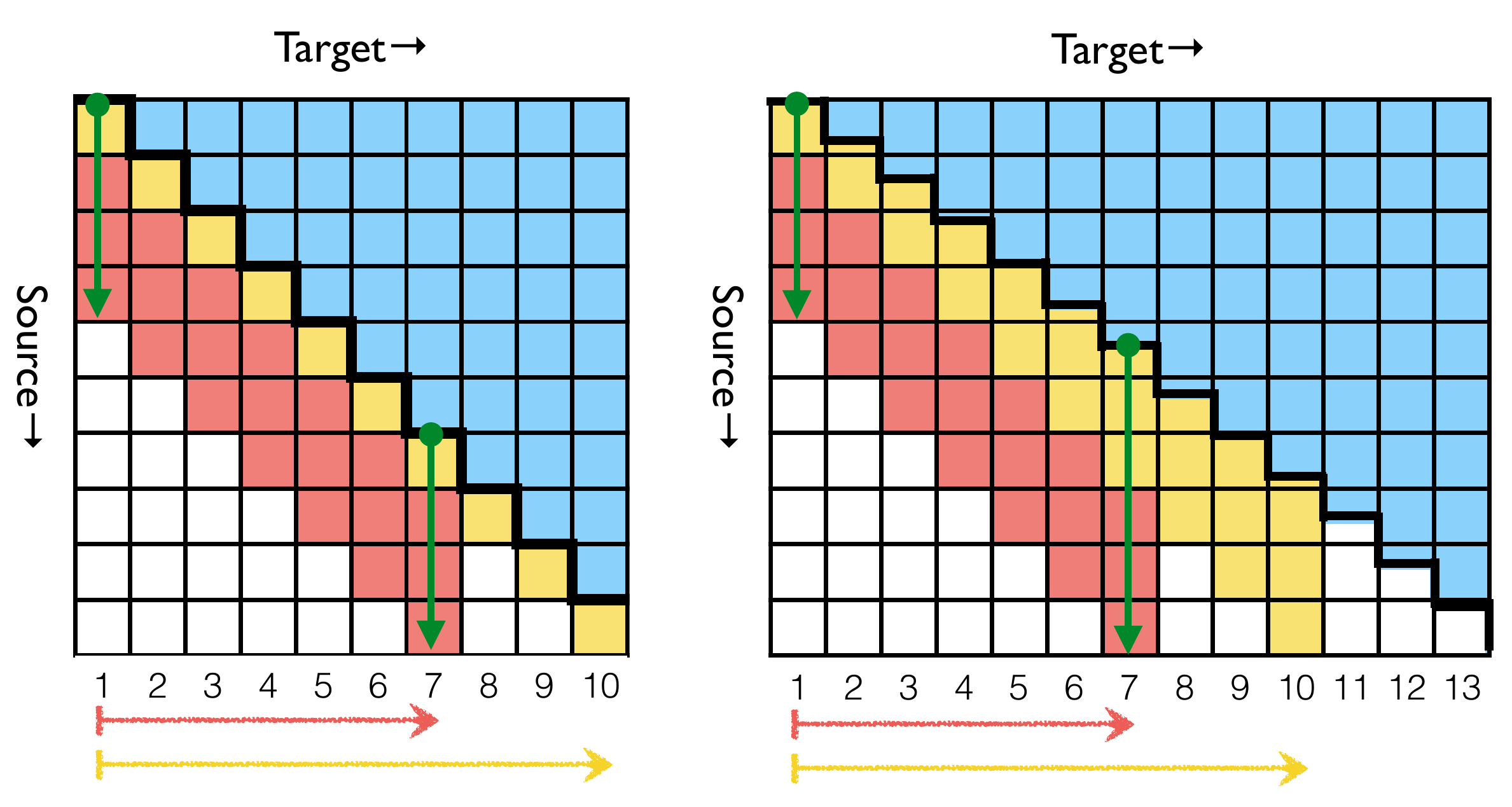}
\captionof{figure}{Illustration of our proposed Average Lagging latency metric. The left figure shows a simple case 
  when $|\vecx|=|\vecy|$ while the right figure shows a more general case when $|\vecx| \neq |\vecy|$.
  The red policy is wait-4,  the yellow is  wait-1, and the thick black is 
  a policy whose AL is 0. 
}
\label{fig:al}
\end{figure}

Inspired by the idea of ``lagging behind the ideal policy'',
we propose a new metric called ``average lagging'' (AL),
shown in Fig.~\ref{fig:al}. 
The goal of AL is to quantify the degree the user is out of sync with the speaker, in terms of the number of source words.
The left figure shows a special case when $|\vecx|=|\vecy|$ for simplicity reasons. 
The thick black line indicates the ``wait-0'' policy where the decoder is alway one word {\em ahead} of the encoder
and we define this policy to have an AL of 0.
The diagonal yellow policy is our ``wait-1'' which is always one word behind the wait-0 policy. 
In this case, we define its AL to be 1.
The red policy is our wait-4, and it is always 4 words behind the wait-0 policy, so its AL is 4.
Note that in both cases, we only count up to (but including) the cut-off point (indicated by the horizontal yellow/red arrows,
or 10 and 7, resp.) because the tail can be generated instantly without  further delay.
More formally, for the ideal case where $|\vecx=|\vecy|$, we can define:
\begin{equation}
\AL_g(\vecx,\vecy) = \frac{1}{\tau_g(|\vecx|)} \sum_{t = 1}^{\tau_g(|\vecx|)} g(t) - (t-1)
\label{eq:AL}
\end{equation}
We can infer that the AL for wait-$k$ is exactly $k$.

When we have more realistic cases like the right side of Fig.~\ref{fig:al} when $|\vecx|<|\vecy|$,
there are more and more delays accumulated when target sentence grows.
For example, for the yellow wait-1 policy has a delay of more than 3 words at decoding its cut-off step 10,
and the red wait-4 policy has a delay of  almost 6 words at its cut-off step 7.
This difference is mainly caused by the tgt/src ratio.
For the right example, there are $1.3$ target words per source word.
More generally, we need to offset the ``wait-0'' policy and redefine: 
\begin{equation}
\AL_g(\vecx,\vecy) = \frac{1}{\tau_g(|\vecx|)} \sum_{t = 1}^{\tau_g(|\vecx|)} g(t) - \frac{t-1}{r}
\label{eq:AL2}
\end{equation}
where $\tau_g(|\vecx|)$ denotes the cut-off step,
and $r=|y|/|x|$ is the target-to-source length ratio.
We observe that wait-$k$ with catchup has an AL  $\simeq k$.


\vspace{-0.1cm}
\section{Implementation Details}
\vspace{-0.1cm}
\label{sec:detail}

While RNN-based implementation of our wait-$k$ model is straightforward
and our initial experiments showed equally strong results,
due to space constraints we will only present Transformer-based results.
Here we describe the implementation details for training a prefix-to-prefix Transformer,
which is a bit more involved than RNN.

\subsection{Background: Full-Sentence Transformer}

We first briefly review the Transformer architecture step by step 
to highlight the difference between the conventional and simultaneous Transformer.
The encoder of Transformer works in a self-attention fashion and takes an input 
sequence $\vecx$,
and produces a new sequence of hidden states $\vecz = (z_1,...,z_n)$ 
where $z_i \in \mathbb{R}^{d_z}$ is as follows:
\begin{equation}
z_i = \textstyle\sum_{j=1}^n \alpha_{ij} \;  P_{W_\text{V}}(x_j)
\label{eq:weightedcontext}
\end{equation}
Here $P_{W_\text{V}}(\cdot)$ is a projection function from the input space to the value space, 
and $\alpha_{ij}$ denotes the attention weights: 
\begin{equation}
  \alpha_{ij}\!=\!\frac{\exp e_{ij}}{\sum_{l=1}^n \exp e_{il}},
  \; e_{ij}\!=\!\frac{P_{W_\text{Q}}(x_i) P_{W_\text{V}}(x_j)^T} {\sqrt{d_x}}
\label{eq:simisoft}
\end{equation}
where $e_{ij}$ measures similarity between inputs. 
Here $P_{W_\text{Q}}(x_i)$ and $P_{W_\text{K}}(x_j)$ project $x_i$ and $x_j$ to query and key spaces, resp.
We use 6 layers of self-attention 
and use $\vech$ to denote the top layer output sequence (i.e., the source context).

On the decoder side, during training time, the  gold output sequence 
$\vecy^* = (y_1^*,...,y_m^*)$ goes 
through the same  self-attention to generate hidden self-attended 
state sequence $\vecc = (c_1,...,c_m)$. 
Note that because decoding is incremental, we let $\alpha_{ij} = 0$ if $j>i$ in Eq.~\ref{eq:simisoft}
to restrict  self-attention to previously generated words.

In each layer, after we gather all the hidden representations for each target word 
through self-attention,
we perform target-to-source attention: 
\begin{equation*}
c'_i = \textstyle\sum_{j=1}^n \beta_{ij} \;  P_{W_{\text{V}'}}(h_j)
\end{equation*}
similar to self-attention, $\beta_{ij}$ measures the similarity between $h_j$ and $c_i$ 
as in Eq.~\ref{eq:simisoft}. 

\subsection{Training Simultaneous  Transformer}

Simultaneous translation requires feeding the source 
words incrementally to the encoder,
but a naive implementation of such
incremental encoder/decoder is inefficient.
Below we describe a faster implementation.

For the encoder, during training time, 
we still feed the entire sentence at once 
to the encoder. 
But different from the self-attention layer 
in conventional Transformer (Eq.~\ref{eq:simisoft}),
we constrain each source word to 
attend to its predecessors only (similar to decoder-side self-attention),
effectively simulating an incremental encoder: 
\begin{equation*}
  \alpha^{(t)}_{ij} = \left\{\begin{matrix}  
  \frac{\exp e^{(t)}_{ij}}{\sum_{l=1}^{g(t)} \exp e^{(t)}_{il}}  & \text{if} \;\; {i,j}  \leq g(t)  \\[0.05cm]
  0 &\text{otherwise}
  \end{matrix}\right.
\label{eq:simulsoft}
\end{equation*}
\begin{equation*}
e^{(t)}_{ij} =  \left\{\begin{matrix} 
\frac{P_{W_\text{Q}}(x_i) \; P_{W_\text{K}}(x_j)^T} {\sqrt{d_x}} & \text{if} \;\; {i,j}  \leq g(t)  \\[0.05cm]
-\infty  & \text{otherwise}
\end{matrix}\right. 
\label{eq:simulweights}
\end{equation*}

Then we have a newly defined hidden state sequence 
$\vecz^{(t)}= (z^{(t)}_1,...,z^{(t)}_n)$ at decoding step $t$:
\begin{equation}
z^{(t)}_i = \textstyle\sum_{j=1}^n \alpha^{(t)}_{ij} \;  P_{W_\text{V}}(x_j)
\label{eq:weightedcontextwitht}
\end{equation}
When a new source word is received,
all previous source words need to adjust their representations.

\section{Experiments}

\begin{figure*}
\begin{tabular}{cc}
\centering
  \includegraphics[width=7.cm]{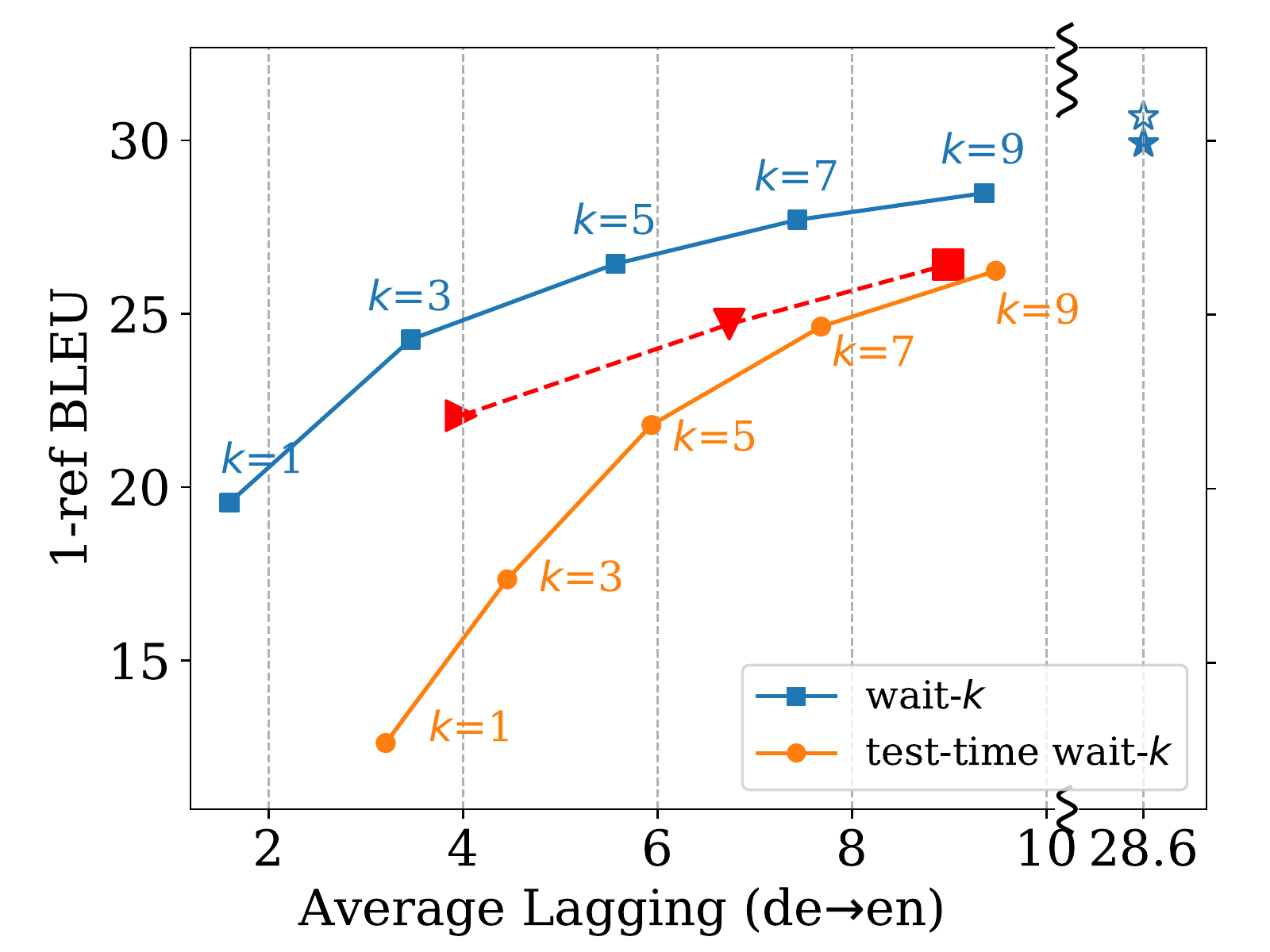} & \centering
  \includegraphics[width=7.cm]{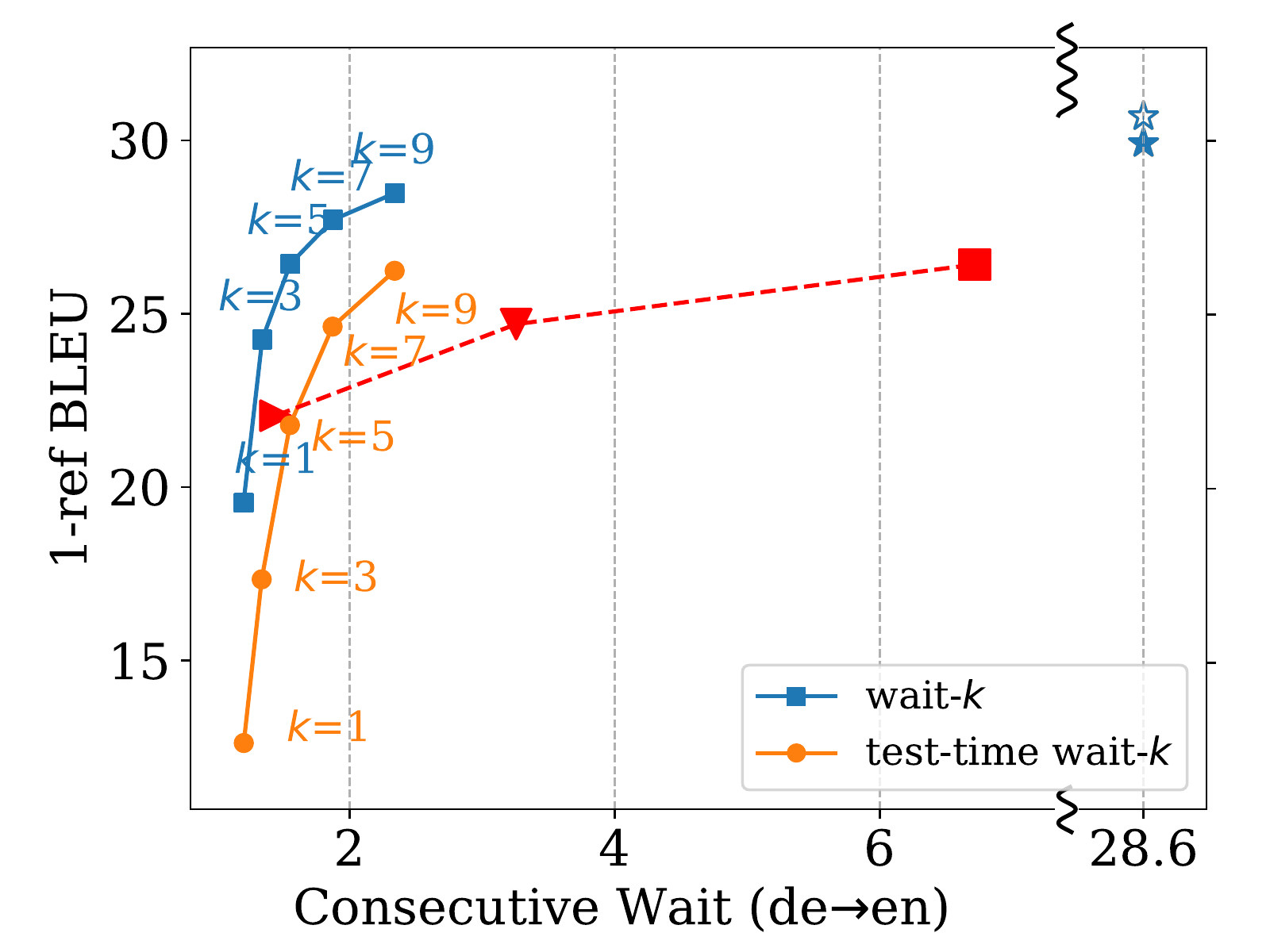}
\end{tabular}\\[-0.4cm]
\captionof{figure}{Translation quality against latency metrics (AL and CW) on German-to-English simultaneous translation,
  showing wait-$k$ 
  and test-time wait-$k$ results, full-sentence baselines, and our adaptation of \namecite{gu+:2017}
 (\textcolor{red}{{$\blacktriangleright$}}:CW=2;
  \textcolor{red}{$\blacktriangledown$}:CW=5;
  \textcolor{red}{$\blacksquare$}:CW=8),
  all based on the same Transformer.
\textcolor{colorC0}{$\bigstar$\ding{73}}:full-sentence (greedy and beam-search).
}
\label{fig:APCW_de2en}
\smallskip

\begin{tabular}{cc}
\centering
  \includegraphics[width=7.cm]{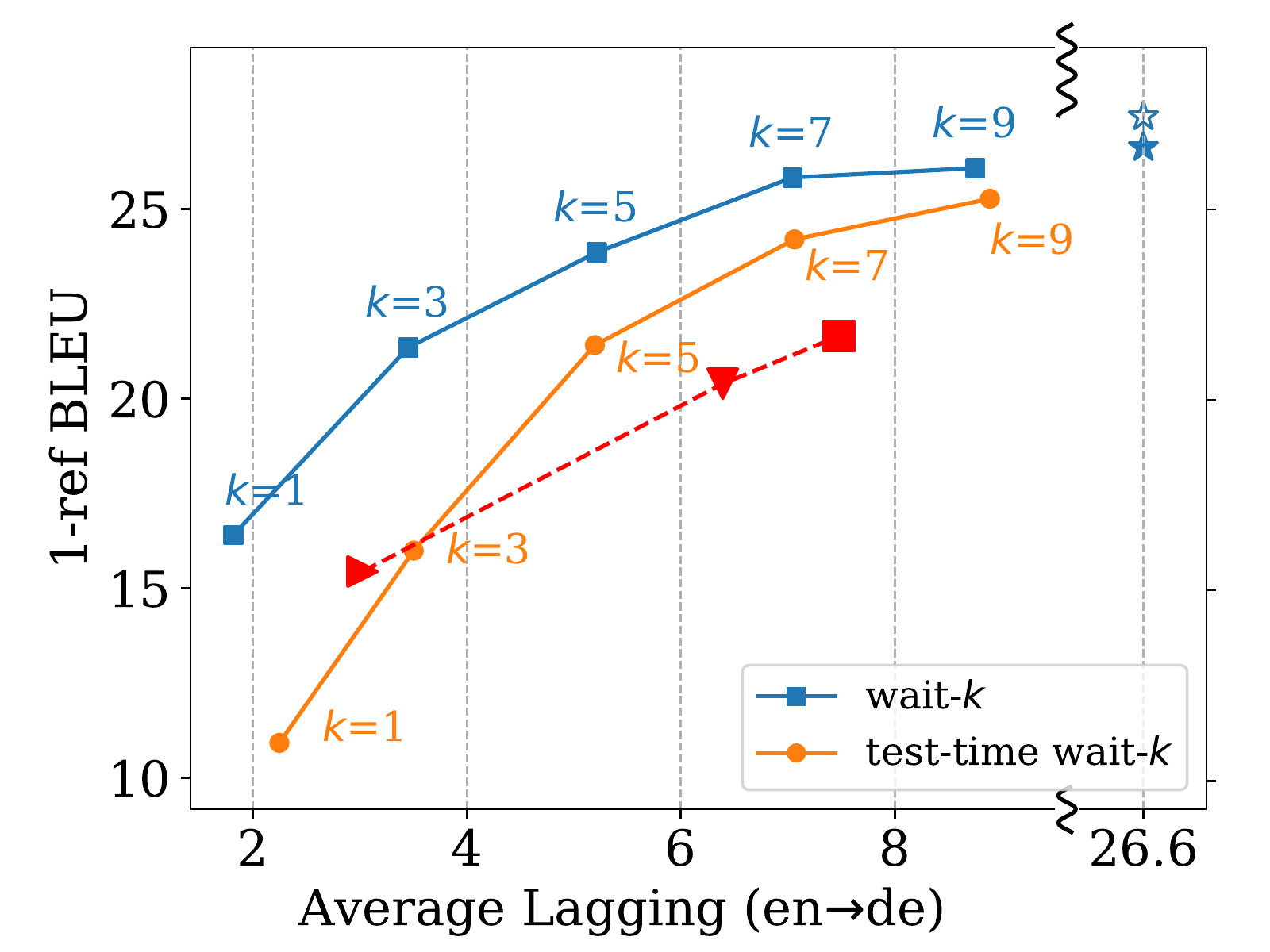} & \centering
  \includegraphics[width=7.cm]{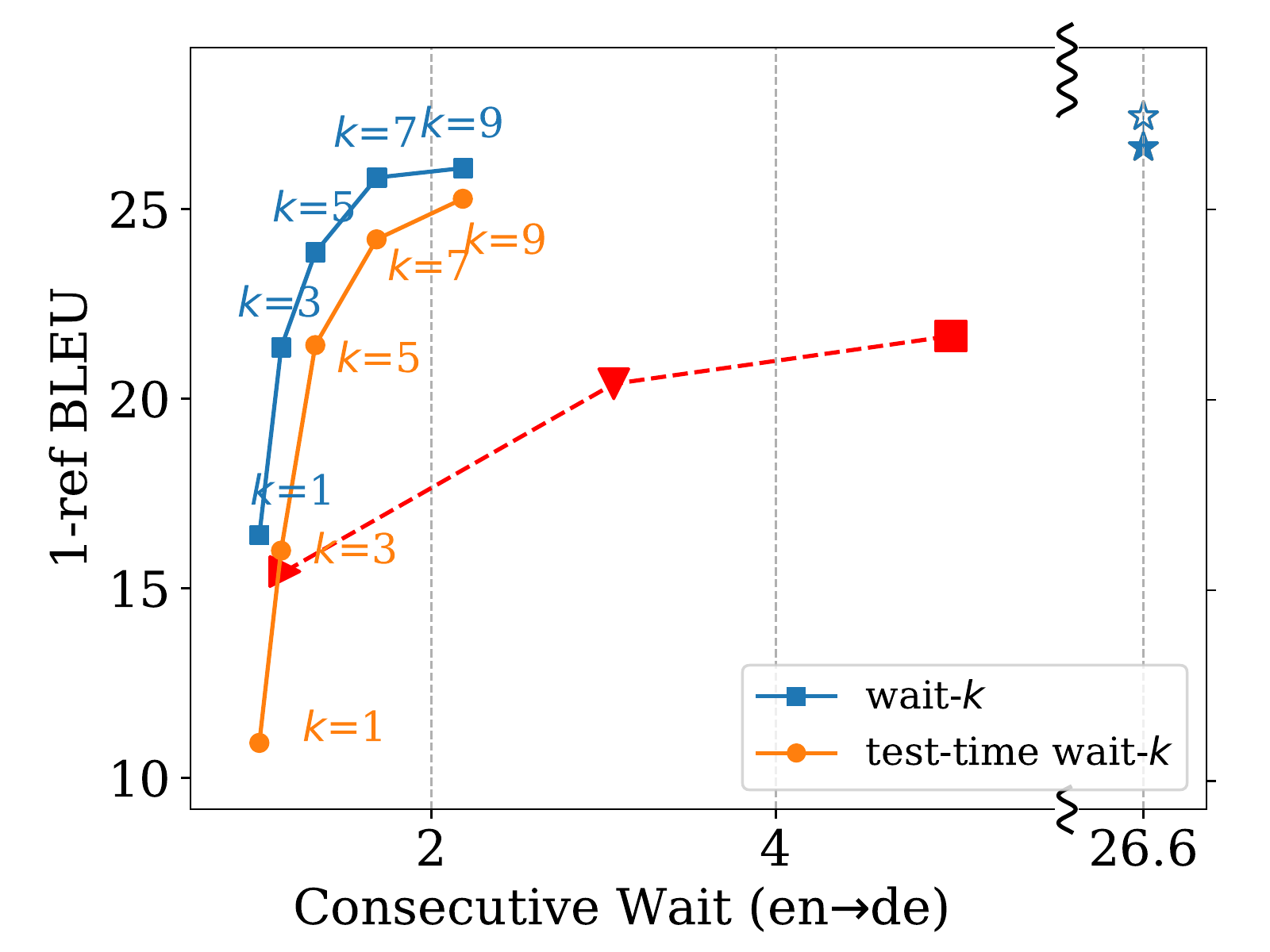}
\end{tabular}\\[-0.4cm]
\captionof{figure}{
  Translation quality against latency metrics on English-to-German simultaneous translation.
}
\label{fig:APCW_en2de}

\smallskip

\begin{tabular}{cc}
\centering
  \includegraphics[width=7.cm]{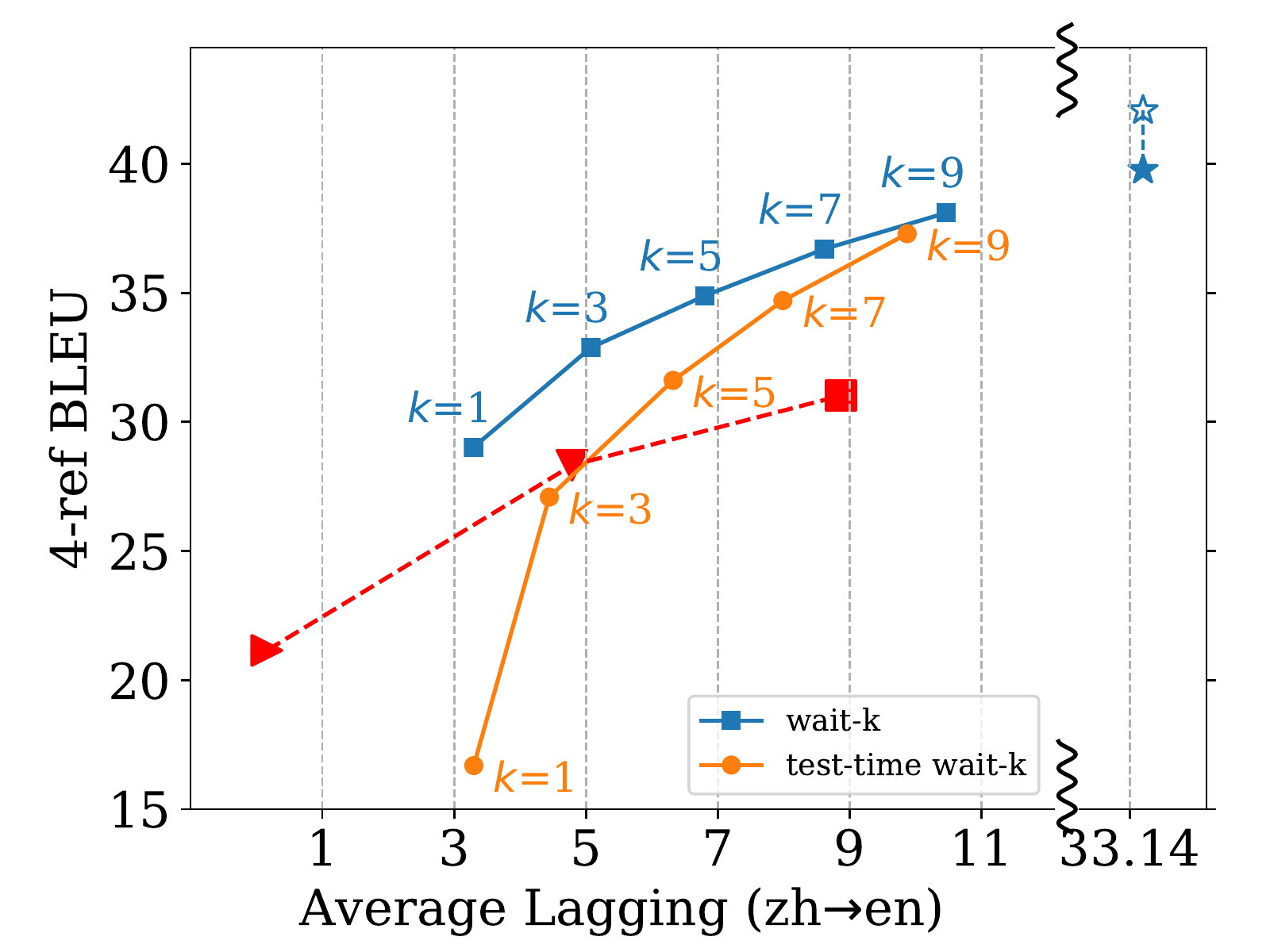} & \centering
  \includegraphics[width=7.cm]{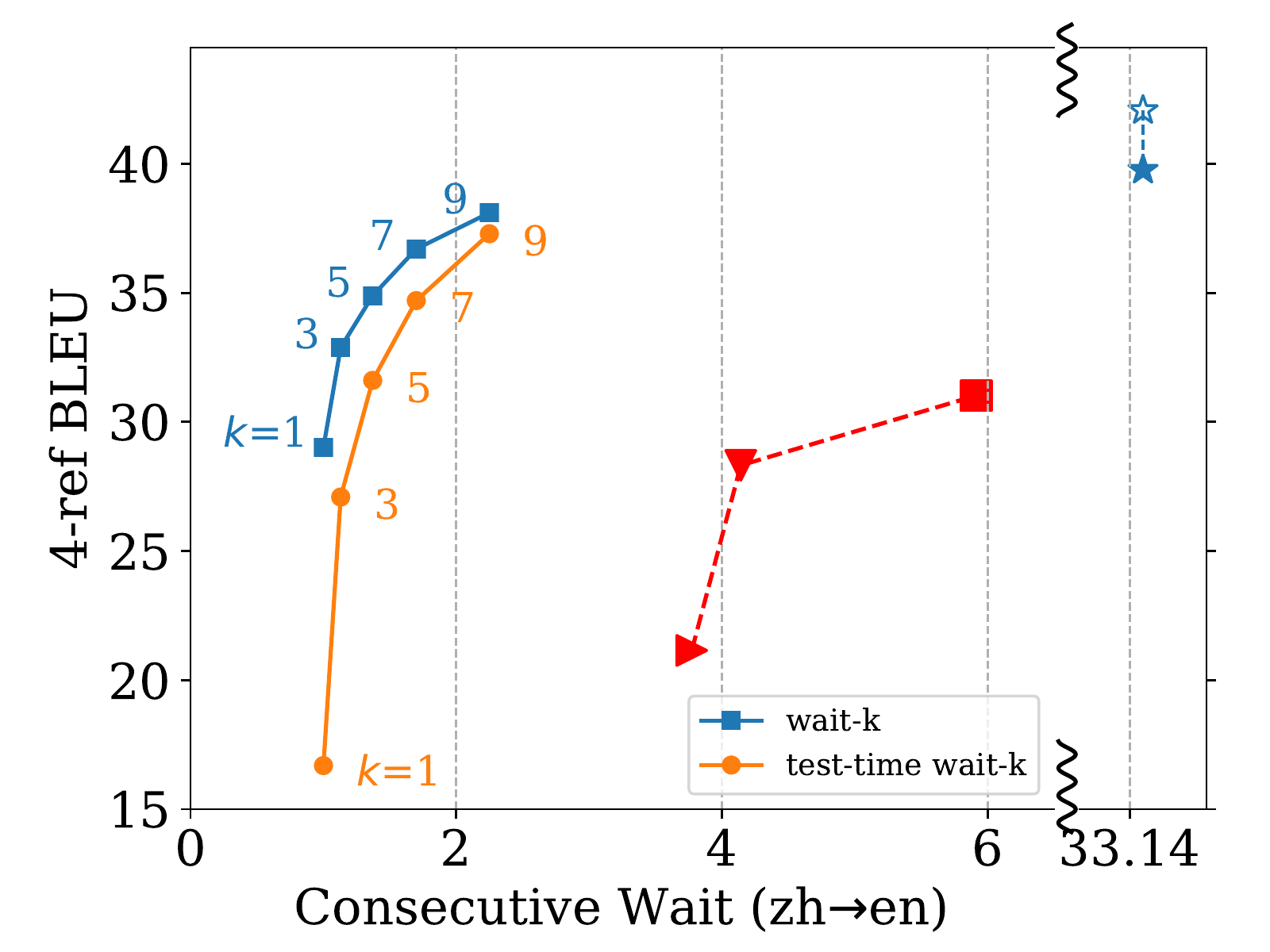}
\end{tabular}\\[-0.4cm]
\captionof{figure}{
  Translation quality against latency on Chinese-to-English simultaneous translation.
}
\label{fig:APCW_zh2en}

\smallskip

\begin{tabular}{cc}
\centering
  \includegraphics[width=7.cm]{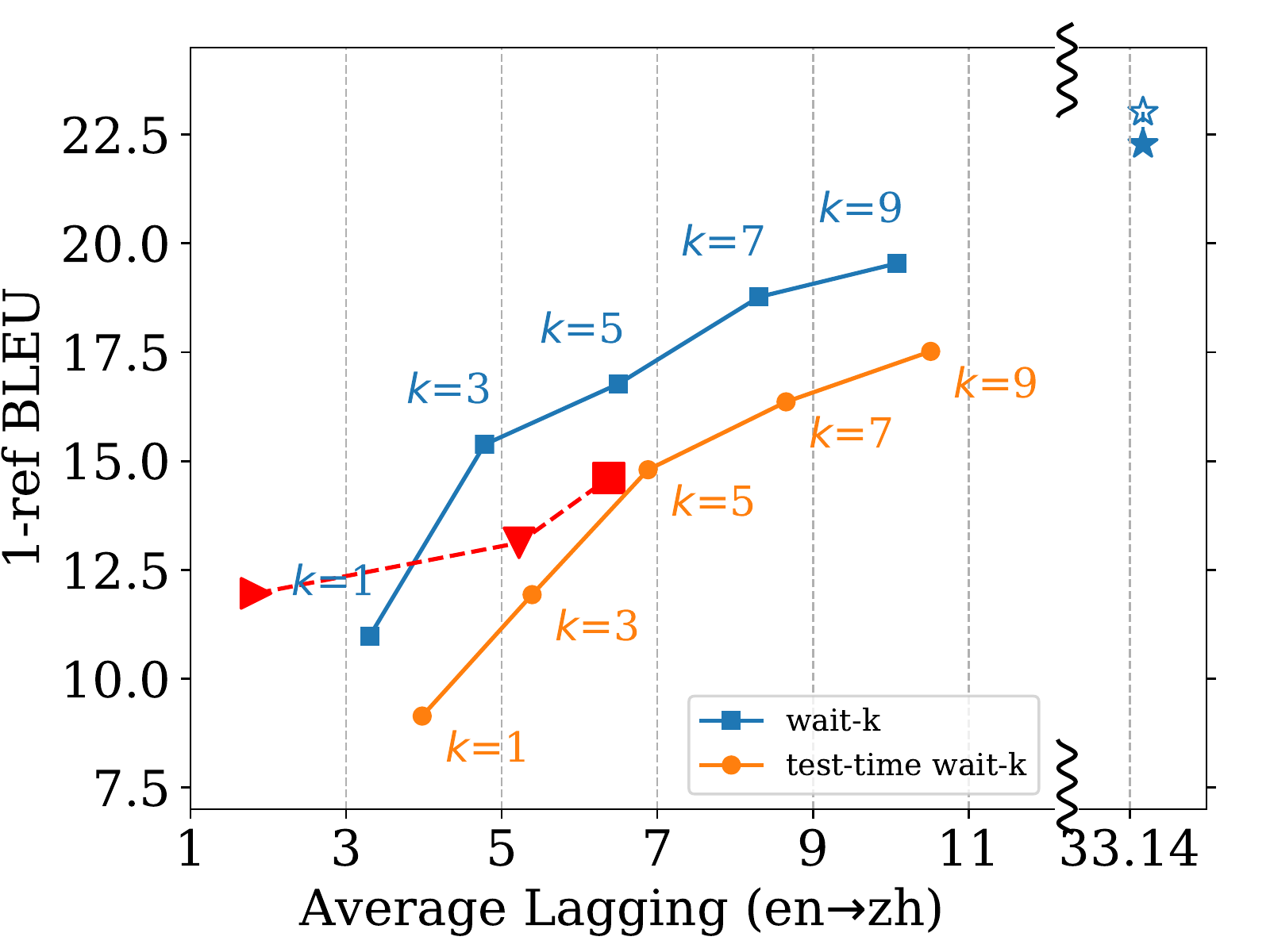} & \centering
  \includegraphics[width=7.cm]{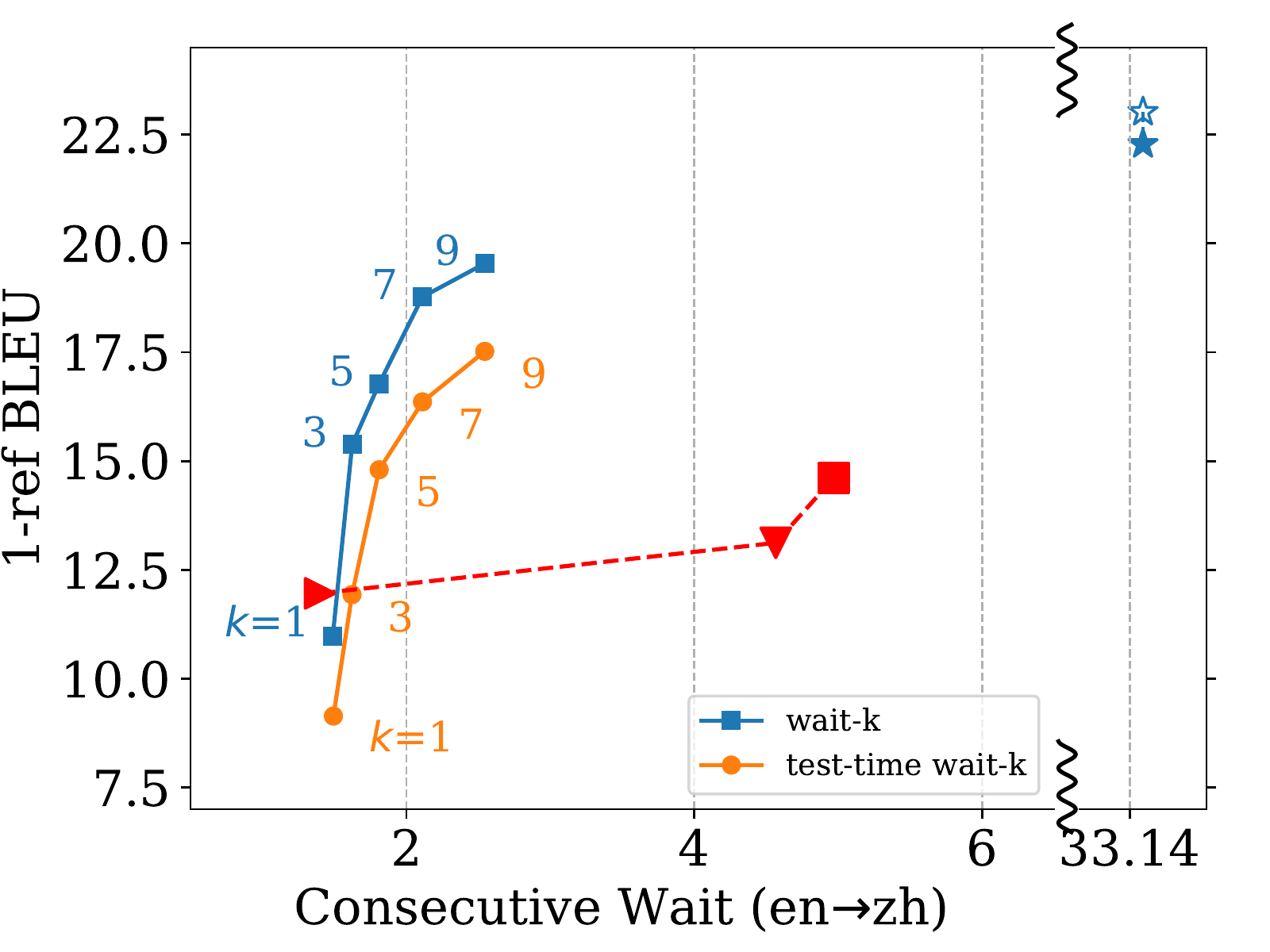}
\end{tabular}\\[-0.4cm]
\captionof{figure}{
  Translation quality against latency  on English-to-Chinese, with encoder catchup (see Appendix~\ref{sec:catchup}). 
}
\label{fig:APCW_en2zh}
\end{figure*}



\vspace{-0.2cm}
\subsection{Datasets and Systems Settings}

We evaluate our work on four simultaneous translation directions:
German$\leftrightarrow$English and Chinese$\leftrightarrow$English.
For the training data, we use the parallel corpora available from WMT15\footnote{\!\!\!\scriptsize\url{http://www.statmt.org/wmt15/translation-task.html}}
for German$\leftrightarrow$English (4.5M sentence pairs) and 
NIST corpus for Chinese$\leftrightarrow$English (2M sentence pairs).
We first apply BPE~\cite{sennrich+:2015} on all texts in order 
to reduce the vocabulary sizes. 
For German$\leftrightarrow$English evaluation, 
we use newstest-2013 (dev) as our dev set and newstest-2015 (test) as our test set,
with 3,000 and 2,169 sentence pairs, respectively.
For Chinese$\leftrightarrow$English evaluation,
we use NIST 2006 and 
NIST 2008 as our dev and test sets.
They contain 616 and 691 Chinese sentences, each with 4 English references.
When translating from Chinese to English, we report 4-reference BLEU scores,
and in the reverse direction, we use the second among the four English references as the source text,
and report 1-reference BLEU scores.



Our implementation is adapted from PyTorch-based OpenNMT \cite{klein+:2017}.
Our Transformer is essentially the same as the base model from 
the original paper \cite{vaswani+:2017}.


\if

Before presenting our experiments, we first explain below the two standard latency metrics for simultaneous translation. 
Consecutive Wait (CW)  \cite{gu+:2017} measures the average lengths of conseuctive wait segments
(the best case is 1 for our wait-1 and the worst case is $|\vecx|$ for full-sentence MT).
For a policy $g(\cdot)$, the per-step CW at step $t$ is
\(
\CW_g(t) = g(t) - g(t-1).
\)
The CW of a sentence-pair $(\vecx, \vecy)$ is the average CW over all consecutive wait segments:
\[
\resizebox{.48\textwidth}{!}{
  $\CW_g(\vecx,\vecy) = \displaystyle\frac{\sum_{t=1}^{|\vecy|} \CW_g(t)}{\sum_{t=1}^{|\vecy|} \mathbbm{1}_{\CW_g(t) > 0}} = \frac{|\vecx|}{\sum_{t=1}^{|\vecy|} \mathbbm{1}_{\CW_g(t) > 0}}$
  }
\]
\vspace{-0.3cm}

Another latency measurement, Average Proportion (AP) \cite{Cho+:16} measures
the proportion of the area above a policy path in Fig.~\ref{fig:idea}:
\vspace{-0.2cm}
\begin{equation}
\AP_g(\vecx,\vecy) = \frac{1}{|\vecx|\;|\vecy|} \textstyle\sum_{t=1}^{|\vecy|} g(t) 
\label{eq:AP}
\end{equation}
\fi

\vspace{-0.3cm}
\subsection{Quality and Latency of Wait-$k$ Model}
\vspace{-0.1cm}

\begin{table}
\centering
\resizebox{0.48\textwidth}{!}{
\begin{tabular}{ | c |c |c |c |c |c |c | }
\hline
\diagbox{\!Train}{\! Test} & $k$=1 & $k$=3 & $k$=5 & $k$=7 & $k$=9 & $k$=$\infty$ \\
\hline
$k'$=1 & {\em 34.1} & 33.3 & 31.8 & 31.2 & 30.0 & 15.4 \\ \hline
$k'$=3 & {\bf 34.7} & 36.7 & {\em 37.1} & 36.7 & 36.7 & 18.3 \\ \hline
$k'$=5 & 30.7 & 36.7 & 37.8 & 38.4 & {\em 38.6} & 22.4 \\ \hline
$k'$=7 & 31.0 & {\bf 37.0} & {\bf 39.4} & {\em 40.0} & 39.8 & 23.7 \\ \hline
$k'$=9 & 26.4 & 35.6 & 39.1 & {\bf 40.1} & {\bf {\em 41.0}} & 28.6 \\ \hline
$k'$=$\infty$ & 21.8 & 30.2 & 36.0 & 38.9 & 39.9 &  {\bf {\em 43.2}}\\ \hline


\end{tabular} 
}
\caption{wait-$k$ policy  in training and test
  (4-ref BLEU, \zhtoen dev set). The bottom row is ``test-time wait-$k$''.
Bold: best in a column; italic: best in a row.}
\label{tb:combo}
\vspace{-.3cm}
\end{table}

Tab.~\ref{tb:combo} shows the results
of a model trained with wait-$k'$
but decoded with wait-$k$ (where $\infty$ means full-sentence).
Our wait-$k$ is the diagonal,
and the last row is the ``test-time wait-$k$'' decoding.
Also, the best results of wait-$k$ decoding is often
from a model trained with a slightly larger $k'$.

\newcolumntype{M}{>{\em}c}

Figs.~\ref{fig:APCW_de2en}--\ref{fig:APCW_en2zh}
plot translation quality (in BLEU) against latency (in AL and CW) for full-sentence baselines, 
our wait-$k$, test-time wait-$k$ (using full-sentence models),
and our adaptation of \namecite{gu+:2017} from RNN to Transformer\footnote{
However, it is worth noting that, despite our best efforts, we  failed to reproduce their work on their original RNN, 
regardless of using their code or our own.
That being said, our successful implementation of their work on Transformer
is also a notable contribution of this work.
By contrast, it is very easy to make wait-$k$ work on either RNN or Transformer.}
on the same Transformer baseline.
In all these figures, we observe that, as $k$ increases, 
(a) wait-$k$ improves in BLEU score and worsens in latency, and
(b) the gap between test-time wait-$k$ and wait-$k$ shrinks.
Eventually, both wait-$k$ and test-time wait-$k$ approaches the full-sentence baseline
as $k\goesto\infty$.
These results are consistent with our intuitions.

We next compare our results with our adaptation of \namecite{gu+:2017}'s two-staged full-sentence model + reinforcement learning on Transformer.
We can see that while on BLEU-vs-AL plots, their models perform similarly to our test-time wait-$k$ for \detoen and \zhtoen, and
slightly better than our test-time wait-$k$ for \entozh,
  which is reasonable as both use a full-sentence model at the very core.
However, on BLEU-vs-CW plots, their models have much worse CWs, which is also consistent with results in their paper (Gu, p.c.).
This is because their R/W model prefers consecutive segments of READs and WRITEs (e.g., their model often produces R R R R R W W W W R R R W W W W R ...) while our wait-$k$ translates concurrently with the input
(the initial segment has length $k$, and all others have length 1, thus a much lower CW).
We also found their training to be extremely brittle due to the use of RL
whereas our work is very robust.

\begin{figure*}

\resizebox{\textwidth}{!}{%
\setlength{\tabcolsep}{.7pt}
\centering
\begin{tabu}{ c | c c c c c c c c c c c c c c c c c c c     l  }
\rowfont{\small}
  & 1 & 2 & 3 & 4 & 5 & 6 & 7 & 8 & 9 & 10 & 11 & 12  & 13 & 14 & 15 & 16 & 17 & 18 & 19  \\
       & doch & w\"{a}hrend & man & \color{purple}sich & im & kongre- & ss &  nicht & auf & ein & \!\!\!vorgehen & \color{blue}einigen & \color{red}kann & , & warten & mehrere & bs. & nicht & l\"{a}nger \\
\rowfont{\small}
  & but & while & they\,  & -self & in  & \multicolumn{2}{c}{congress} & not & on & one & action & \color{blue}agree & \color{red}can & , & wait & several  & states & no & longer \\
  \hline
  $k$=3  &     &       &  & but & , & while & congress & \color{red}has & not & \,\color{blue}agreed & on & a & course & of & action & , & several & states & no & longer wait \\
\end{tabu}
}
\caption{German-to-English example in the dev set with anticipation.
  The main verb in the embedded clause, ``einigen'' (agree), is correctly predicted 3 words ahead of time (with ``sich'' providing a strong hint),
  while the aux.~verb ``kann'' (can) is predicted as ``has''.
  The baseline translation is ``but , while congressional action can not be agreed , several states are no longer waiting''.
  bs.: bunndesstaaten. 
}
\label{fig:congress}
\smallskip

\resizebox{.9\textwidth}{!}{%
\setlength{\tabcolsep}{1pt}
\centering
\begin{tabu}{ c | c c c c c c c c c c c c l l  }
\rowfont{\small}
 & 1 & 2 & 3 & 4 & 5 & 6 & 7 & 8 & 9 & 10 & 11 & 12    \\[-0.1cm]
  \rowfont{\small}
& \chn{t\=a} &\chn{h\'ai} & \chn{shu\=o} & \chn{xi\`anz\`ai} & \chn{zh\`engz\`ai} & \chn{w\`ei} & \chn{zh\`e} & \chn{y\=\i} & \chn{f\v{a}ngw\`en} & \color{purple}\chn{zu\`o} & \chn{ch\=u} & \color{blue}\chn{\=anp\'ai} \\

  & 他 & 还&  说 &现在& 正在& 为 &这 &一 &访问 &\color{purple}作 &出  &\color{blue}安排\\ 

 \rowfont{\small}
 & he & also & said & now  & (\chn{prog.})$^\diamond$ & for & this & one & visit & \color{purple}make & out & \!\!\!\!\color{blue}arrangement\!\!\!\!\\  
 \hline
 $k$=1 & & he & also& said & that & he & is & now & \color{purple}making & \color{blue}preparations & for  & this  & visit \\
 $k$=3 &  & & & he &  also & said & that & he & is & \color{purple}making & \color{blue}preparations & for  & this  visit\\
 \hline
$k$=$\infty$ & \multicolumn{12}{l}{} & he also said that arrangements  \\
& \multicolumn{12}{l}{} & are being made for this visit 
\end{tabu} 
}
\caption{Chinese-to-English example in the dev set with anticipation.
  Both wait-1 and wait-3 policies yield perfect translations,
  with ``making preparations'' predicted well ahead of time.
  $^\diamond$: progressive aspect marker.
}

 \smallskip


\resizebox{0.9\textwidth}{!}{%
\setlength{\tabcolsep}{1pt}
\centering
\begin{tabu}{ c | c c c c c c c c c c c l }
\rowfont{\small}
 & 1 & 2 & 3 & 4 & 5 & 6 & 7 & 8 & 9 & 10 & 11 &   \\[-0.1cm]
  \rowfont{\small}
 &\chn{jiāng} & \chn{zémín} & \color{purple}\chn{duì} & \color{purple}\chn{bùshí} & \color{purple}\chn{zǒngtǒng} & \color{colorC1}\chn{lái} & \color{colorC1}\chn{huá} & \color{cyan}\chn{fǎngwèn} & \color{blue}\chn{biǎoshì} & \color{brown}\chn{rèliè} & \color{brown}\chn{huānyíng}\\
 & 江 & 泽民 & \color{purple}对 & \color{purple}布什 & \color{purple}总统 & \color{colorC1}来 & \color{colorC1}华 & \color{cyan}访问 & \color{blue}表示 & \color{brown}热烈 & \color{brown}欢迎\\ [-0.1cm]
 \rowfont{\small}
 & jiang & zeming & \color{purple}to & \color{purple}bush & \color{purple}president & \color{colorC1}come-to & \color{colorC1}china & \color{cyan}visit & \color{blue}express & \color{brown}warm & \color{brown}welcome \\ 
 \hline
 $k$=3 &&&&jiang & zemin & \color{blue}expressed & \color{brown}welcome & \color{purple}to & \color{purple}president & \color{purple}bush & \color{purple}'s & \color{cyan}visit \color{colorC1}to china \\
\hline
 $k$=3$^\dagger$ &&&& jiang & zemin & \color{red}meets & \color{purple}president & \color{purple}bush & \color{colorC1}in & \color{colorC1}china & 's & bid to \color{cyan}visit \color{colorC1}china \\
\end{tabu} 
}
\caption{Chinese-to-English example from online news. 
Our wait-3 model correctly
anticipates both ``{\color{blue}expressed}" and ``{\color{brown}welcome}'' (though missing ``{\color{brown}warm}''),
and moves the PP (``{\color{purple}to} ... {\color{cyan}visit} {\color{colorC1}to china}") to the very end which is fluent in the English word order.
$^\dagger$: test-time wait-$k$
produces nonsense translation.
}
  \smallskip
  
\resizebox{0.9\textwidth}{!}{
\centering
\setlength{\tabcolsep}{2pt}
\begin{tabu}{c | c c c c c c c c c c l l }
\rowfont{\small}
 & 1 & 2 & 3 & 4 & 5 & 6 & 7 & 8 & 9 & 10 \\[-0.1cm]
 \rowfont{\small}
 & \meiguo & \dangju & \dui & \shate & \jiezhe & \shizong & \yi & \an &\color{cyan}\gandao &  \color{blue}\danyou \\
(a) & 美国 &  当局 &对 &沙特 &记者 &失踪& 一& 案
 & \color{cyan}感到 &\color{blue}担忧\\
\rowfont{\small}
 \footnotesize & US & authorities & to & Saudi & reporter & \color{purple}missing & a & case & \color{cyan}feel & \color{blue}concern\\
 \hline
 $k$=3 & & & & the & us & authorities & \color{cyan}are & \color{blue}very  & \color{blue}concerned & about & the  saudi  reporter  's  missing  case\\
\hline
 $k$=3$^\dagger$ & & & & the & us & authorities & \color{red}have & \color{red}dis-  & \color{red}appeared & from & saudi reporters \\
\hline
\rowfont{\small}
     &         &           &               &         &       &    &     &        &     & \color
     {red}\buman\\
(b) & 美国 &  当局 &对 &沙特 &记者 &失踪& 一& 案 & \color{cyan}感到 &\color{red}\bf 不满\\
 \hline
 $k$=3 & & & & the & us & authorities & \color{cyan}are & \color{blue}very  & \color{blue}concerned & about & the  saudi  reporter  's  missing  case\\
 $k$=5 & & & & & & the & us & \!\!\! authorities \!\!\!\!\! & \color{cyan}have & \color{cyan}expressed & {\color{red}\bf dissatisfaction}  with the incident \\
 & \multicolumn{10}{c}{} &   of saudi arabia 's missing reporters\\
\hline
\end{tabu}
}
\caption{(a) Chinese-to-English example from more recent news, clearly outside of 
  our data. Both the verb ``\gandao'' (``feel'') and the predicative ``\danyou'' (``concerned'') are correctly anticipated, probably hinted by ``missing''.
(b) If we change the latter to \buman (``dissatisfied''),
the wait-3 result remains the same (which is wrong) while wait-5
translates conservatively without anticipation.
$^\dagger$: test-time wait-$k$ 
produces nonsense translation.
}
\label{fig:reporter}
\end{figure*}

\begin{figure*}
\resizebox{\textwidth}{!}{%
\setlength{\tabcolsep}{1pt}
\centering
\begin{tabu}{ c | c c c c c c c c c c c c c c c c c c c c c c c l }
\rowfont{\small}
  & 1 & 2 & 3 & 4 & 5 & 6 & 7 & 8 & 9 & 10 & 11 & 12 & 13 &14 & 15 & 16 & 17 & 18 & 19 & 20 & 21 & 22 & 23  \\[-0.1cm]
 & it & was & learned & that & this & is & \color{purple}the & \color{purple}largest & \color{colorC1}fire & \color{colorC1}accident & \color{cyan}in & \color{cyan}the & \color{cyan}medical & \color{cyan}and& \color{cyan}health & \color{cyan}system& \color{brown}nationwide & \color{blue}since & \color{blue}the & \color{blue}founding & \color{blue}of \color{blue}& \color{blue}new & \color{blue}china \\
 \hline
 \begin{tabu}{c}
 \rowfont{\small}\\ \rowfont{\small}\\[-0.1cm]$k$=3 \\[-0.1cm] \rowfont{\small}\\ 
 \hline
 \rowfont{\small}\\ $k$=3$^\dagger$ \\[-0.1cm]  \rowfont{\small}\\
 \end{tabu}
 & 
 \begin{tabu}{c}
 \rowfont{\small}~1~\\ \rowfont{\small}\\ [-0.1cm]\\[-0.1cm]\rowfont{\small}\\\hline
 \end{tabu}
 & 
 \begin{tabu}{c}
 \rowfont{\small}~~~2~~~\\ \rowfont{\small}\\ [-0.1cm]\\[-0.1cm]\rowfont{\small}\\\hline
 \end{tabu}
 & 
 \begin{tabu}{c}
 \rowfont{\small}~~~~~~3~~~~~~~\\ \rowfont{\small}\\ [-0.1cm]\\[-0.1cm]\rowfont{\small}\\\hline
 \end{tabu}
 &
 \multicolumn{21}{l}{
 \begin{tabu}{c c c c c c c c c c c c c c c c c l}
 \rowfont{\small}
 4 & 5 & 6 & 7 & 8 & 9 & 10 & 11 & 12 & 13 & 14 & 15 & 16 & 17 & 18 & 19 & 20\\[-0.1cm]
 \rowfont{\small}
  \chn{jù} & \chn{liǎojiě} & \chn{,} & \chn{zhè} & \chn{shì} & \color{PaleGreen}\chn{zhōngguó} & \color{red}\chn{jìn} & \color{red}\chn{jǐ} & \color{red}\chn{nián} & \color{red}\chn{lái} & \chn{fāshēng} & \chn{de} & \color{purple}\chn{zuì} & \color{purple}\chn{dà} & \color{purple}\chn{yī} & \color{purple}\chn{qǐ} &\color{cyan}\chn{yīliáo} & \chn{\color{cyan}wèishēng xìtǒng \color{colorC1}~huǒzāi shìgù} \\
  据 & 了解 & , & 这 & 是 & \color{PaleGreen}中国 & \color{red}近 & \color{red}几 & \color{red}年 & \color{red}来 & 发生 & 的 & \color{purple}最 & \color{purple}大 & \color{purple}一 & \color{purple}起 & \color{cyan}医疗 &\color{cyan}~~卫生 ~~系统 \color{colorC1}~火灾  事故 \\[-0.1cm]
\rowfont{\small}
  to & known & ,& this & is & \color{PaleGreen}China & \color{red}recent & \color{red}few & \color{red}years & \color{red}since & happen & - & \color{purple}most & \color{purple}big & \color{purple}one & \color{purple}case & \color{cyan}medical & \color{cyan}~~~health  ~~system \color{colorC1}~~fire accident \\
  \hline
 \rowfont{\small}
  \chn{yīnwèi} & \chn{tā} & \chn{shì} & \chn{,} & \chn{zhègè} & \chn{,} & \chn{shì} & \color{purple}\chn{zùi} & \color{purple}\chn{dà} & \chn{de} & \color{colorC1}\chn{huǒzāi} & \color{colorC1}\chn{shìgù} & \chn{,} & \color{blue}\chn{zhè} & \color{blue}\chn{shì} & \color{blue}\chn{xīn} & \color{blue}\chn{zhōngguó} & \color{blue}\chn{chénglì} \chn{yǐlái} \\
  因为 & 它 & 是 & , & 这个 & , & 是 & \color{purple}最 & \color{purple}大 & 的 & \color{colorC1}火灾 & \color{colorC1}事故 & , & \color{blue}这 & \color{blue}是 & \color{blue}新 &\color{blue}中国 &\color{blue}~成立 以来 \\[-0.1cm]
  \rowfont{\small}
  because & it & is & , & this & , & is & \color{purple}most & \color{purple}big & - & \color{colorC1}fire & \color{colorC1}accident &,& \color{blue}this & \color{blue}is & \color{blue}new & \color{blue}China &\color{blue}funding since\\
 
 \end{tabu}
 }\\
 \hline
\end{tabu} 
}
\caption{English-to-Chinese example in the dev set with incorrect anticipation due to mandatory long-distance reorderings. The English sentence-final  clause {\color{blue}``since the founding of new china''} is incorrectly predicted in Chinese as {\color{red}``近 几 年 来''(``in recent years'')}. 
Test-time wait-3 produces  translation in the English word order, which
sounds odd in Chinese,
and misses two other quantifiers ({\color{cyan}``in the medical and health system''} and {\color{brown}``nationwide''}), though without prediction errors.
The full-sentence translation, 
``据 了解，这 是 {\color{blue}{新 中国 成立 以来}} ，{\color{brown}全国} {\color{cyan}医疗 卫生 系统} 发生 的 {\color{purple}最 大 的 一 起} {\color{colorC1}火 灾 事故}'', is perfect.}
\label{fig:fire}
\end{figure*}

\vspace{-0.2cm}

\subsection{Human Evaluation on Anticipation}
\vspace{-0.1cm}

\begin{table}
\centering
\resizebox{0.48\textwidth}{!}{
\begin{tabular}{ | l |r |r |r |r |r |r |}
\hline
\multirow{2}{*}{} & $k$=3 & $k$=5 & $k$=7 & $k$=3 & $k$=5 & $k$=7 \\
\cline{2-7}
& \multicolumn{3}{c|}{\color{red} \zhtoen} & \multicolumn{3}{c|}{\color{red} \entozh} \\
\hline
sent-level \% & 33 & 21 & 9 & 52 & 27 & 17 \\
\hline
word-level \% & 2.5 & 1.5 & 0.6 & 5.8 & 3.4 & 1.4 \\
\hfill accuracy & 55.4 & 56.3 & 66.7 &  18.6 & 20.9 & 22.2 \\
\hline
 & \multicolumn{3}{c|}{\color{red} \detoen} & \multicolumn{3}{c|}{\color{red} \entode} \\
\hline
sent-level \% & 44 & 27 & 8 & 28 & 2 & 0 \\
\hline
word-level \% & 4.5 & 1.5 & 0.6 & 1.4 & 0.1 & 0.0 \\
\hfill accuracy & 26.0 & 56.0 & 60.0 & 10.7 & 50.0 & n/a \\
\hline
\end{tabular} 
}
\caption{Human evaluation for all four directions (100 examples each from dev sets). We report sentence- and word-level anticipation rates, and
   the word-level anticipation accuracy (among anticipated words).}
\label{tb:human}
\vspace{-.3cm}
\end{table}

Tab.~\ref{tb:human} shows human evaluations on anticipation rates and  accuracy on all four directions, using 100 examples in each language pair from the dev sets.
As expected, we can see that, with increasing $k$,
the anticipation rates decrease (at both sentence and word levels),
and the anticipation accuracy improves.
Moreover, the anticipation rates are very different among the four directions,
with
\[
\text{\entozh \;  $>$ \; \detoen \; $>$ \; \zhtoen \;  $>$ \; \entode}
\]
Interestingly, this order is exactly the same
with the order of the BLEU-score gaps between our wait-9 and full-sentence models:
\[
\small
\text{\entozh: 2.7 $>$  \detoen: 1.1 $>$ \zhtoen: 1.6$^\dagger >$  \entode: 0.3}
\]
($^\dagger$: difference in 4-ref BLEUs, which in our experience reduces by about half in 1-ref BLEUs).
We argue that this order roughly characterizes the relative
difficulty of {\em simultaneous} translation in these directions.
In our data, we found \entozh to be particularly difficult due to the mandatory long-distance reorderings
of English sentence-final temporal clauses (such as ``in recent years'') to much earlier positions in Chinese;
see Fig.~\ref{fig:fire} for an example.
It is also well-known that \detoen is  more challenging in simultaneous translation than \entode since SOV$\rightarrow$SVO involves prediction of the verb,
while SVO$\rightarrow$SOV generally does not need prediction in our wait-$k$ with a reasonable $k$,
because V is often shorter than O.
For example, human evaluation found only 1.4\%, 0.1\%, and 0\% word anticipations in \entode for $k$=3, 5 and 7, 
and 4.5\%, 1.5\%, and 0.6\% for \detoen.

\vspace{-0.2cm}

\subsection{Examples and Discussion}
\vspace{-0.1cm}

We showcase some examples in \detoen and \zhtoen
from the dev sets and online news in Figs.~\ref{fig:congress} to~\ref{fig:reporter}.
In all these examples except Fig.~\ref{fig:reporter}(b),
our wait-$k$ models can generally anticipate correctly,
often producing translations as good as the full-sentence baseline.
In Fig.~\ref{fig:reporter}(b), when we change the last word,
the wait-3 translation remains unchanged (correct for (a) but wrong for (b)),
but wait-5 is more conservative and produces the correct translation without anticipation.

Fig.~\ref{fig:fire} demonstrates a major limitation of our fixed wait-$k$ policies,
that is, sometimes it is just impossible to predict correctly and
you have to wait for more source words. In this example,
due to the required long-distance reordering between English and Chinese
(the sentence-final English clause has to be placed very early in Chinese),
any wait-$k$ model would not work, and a good policy should wait till the very end.

\vspace{-0.1cm}
\section{Related Work}
\vspace{-0.1cm}

The work of \namecite{gu+:2017} is different from ours in four (4) key aspects:
(a) by design, their model does not anticipate;
(b) their model can not achieve any specified latency metric at test time while our wait-$k$ model is guaranteed to have a $k$-word latency; 
(c) their model is a combination of two models, using a full-sentence base model to translate, thus a mismatch between training and testing,
while our work is a genuine simultaneous model,
and
(d) their training is also two-staged, using RL to update the R/W model, while we train from scratch.

In a parallel work, 
\namecite{press+smith:2018} propose an ``eager translation" model which also outputs target-side words before the whole input sentence is fed in, 
but there are several crucial differences:
(a) their work still aims to translate full sentences using beam search, and is therefore, as the authors admit, ``not a simultaneous translation model'';
(b) their work does not anticipate future words; and
(c) they use word alignments to learn the reordering and achieve it 
in decoding by emitting the $\epsilon$ token, while our work integrates reordering into a single wait-$k$ prediction model that is agnostic of, yet capable of, reordering.

In another recent work, \namecite{ashkan+:2018}
adds a prediction action to the work of \namecite{gu+:2017}.
Unlike \namecite{grissom+:2014} who predict the source verb which might come after several words, 
they instead predict the {\em immediate} next source words, which we argue is not as useful  in SOV-to-SVO translation.
\footnote{
Their codebase on Github is not runnable,
and their baseline 
 is inconsistent with \namecite{gu+:2017} which we compared to,
so we did not include their results for comparison.
}
In any case, we are the first to predict directly {\em on the target side}, thus
integrating anticipation in a single translation model.

\namecite{jaitly+:2016} propose an online neural transducer for speech recognition
that is conditioned on prefixes. This problem does not have reorderings
and thus no anticipation is needed.

\vspace{-0.1cm}
\section{Conclusions} 
\vspace{-0.1cm}

We have presented a  prefix-to-prefix training and decoding framework for
simultaneous translation with integrated anticipation, and a wait-$k$ policy
that can achieve arbitrary word-level latency while
maintaining high translation quality.
This prefix-to-prefix architecture has the potential to be used in other sequence tasks outside of MT
that involve simultaneity or incrementality.
We leave many open questions to future work, e.g.,
adaptive policy using a single model \cite{zheng+:2019}.
\iftrue
{
\section*{Acknowledgments}
{
\vspace{-0.2cm}
  We thank Colin Cherry (Google Montreal)
  for spotting a mistake in  AL (Eq.~\ref{eq:AL}),
  Hao Zhang (Google NYC) for comments,
  the bilingual 
  speakers for human evaluations,
  and the anonymous reviewers for suggestions.
}
}
\fi


\bibliography{main}
\bibliographystyle{acl_natbib} 

\clearpage
\section*{\Large Appendix}
\appendix

\section{Supplemental Material: Model Refinement with Catchup Policy}
As mentioned in Sec.~\ref{sec:pred},
the wait-$k$ decoding 
is always $k$ words behind the incoming source stream. 
In the ideal case where the input and output sentences have equal length,
the translation will finish $k$ steps after the source sentence finishes, i.e., the tail length is also $k$.
This is consistent with human interpreters who start and stop a few seconds after
the speaker starts and stops.

\begin{figure}[t]
\centering
\includegraphics[width=\linewidth]{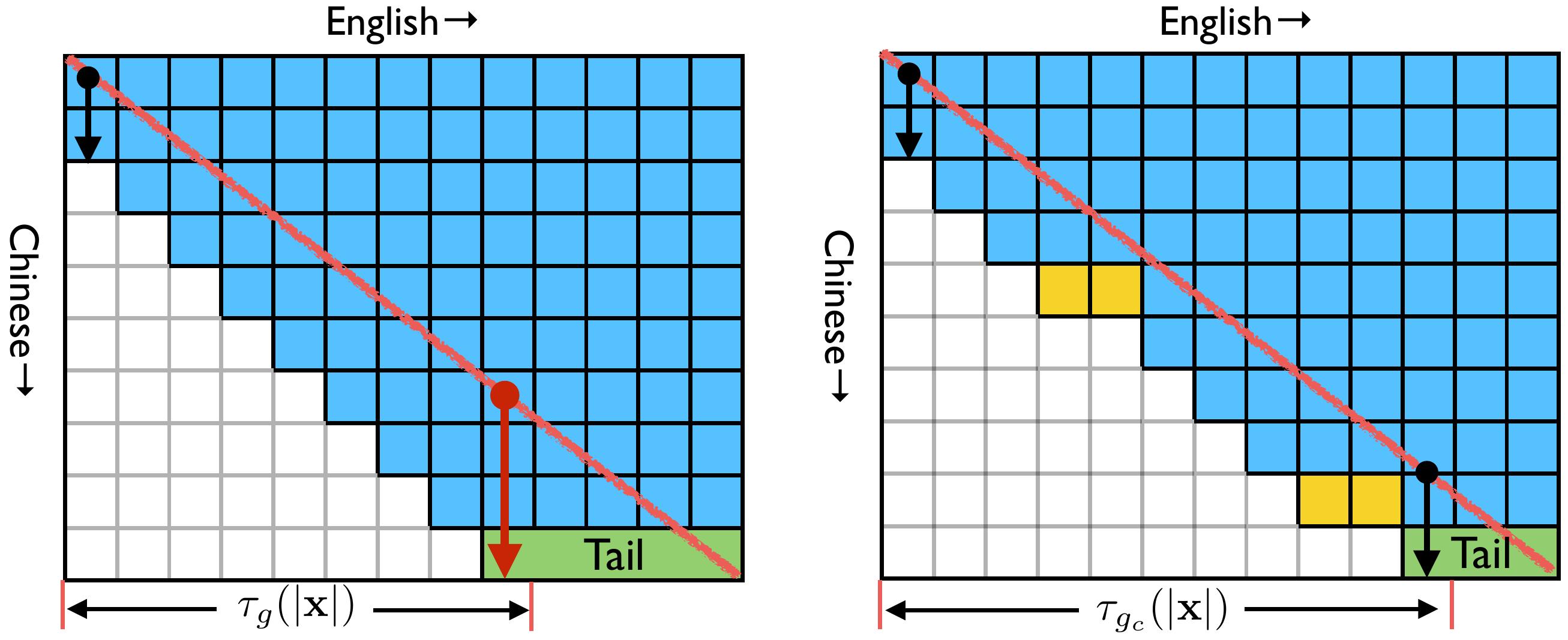}
\captionof{figure}{Left (wait-2): it renders the user increasingly out of sync with the speaker (the diagonal line denotes the ideal perfect synchronization).
Right (+catchup): it shrinks the tail and is closer to the ideal diagonal, reducing the effective latency. 
Black and red arrows illustrate 2 and 4 words lagging behind the diagonal, resp.
}
\label{fig:catchup}
\vspace{-0.5cm}
\end{figure}

However, 
input and output sentences
generally have different lengths.
In some extreme directions such as Chinese to English,
the target side is significantly longer than the source side,
with an average gold tgt/src ratio,
\(r=\overline{|\vecy^\star|/|\vecx|}\),
of around 1.25 \cite{huang+:2017,yilin+:2018}.
In this case, if we still follow the vanilla wait-$k$ policy,
the tail length will be $0.25 |\vecx| + k$ which increases with input length.
For example, given a 20-word Chinese input sentence, the tail of wait-3 policy will be 8 word long, almost half of the source length.
This brings two negative effects:
(a) as decoding progresses, the user will be effectively lagging behind further and further (becomes each Chinese word in principle translates to 1.25 English words),
rendering the user more and more out of sync with the speaker;
and
(b) when a source sentence finishes, the rather long tail is displayed immediately, causing a cognitive burden on the user.\footnote{
It is true that the tail can in principle be displayed concurrently with the first $k$  words of the next input,
but the tail is now much longer than $k$.}
These problems become worse with longer input sentences (see Fig.~\ref{fig:catchup}). 

To address this problem, we devise a ``wait-$k$+catchup'' policy 
so that the user is still $k$ word behind the input in terms of real information content,
i.e., always $k$ source words behind the ideal perfect synchronization policy denoted by the diagonal line in Fig.~\ref{fig:catchup}.
For example, assume the tgt/src ratio is $r=1.25$, 
we will output 5 target words for every 4 source words;
i.e., the catchup frequency, denoted $c=r-1$, is 0.25.
See Fig.~\ref{fig:catchup}.

More formally, with catchup frequency $c$, the new policy is: 
\begin{equation}
\gcatchup(t) = \min\{k+t-1-\floor{ct},\; |\vecx|\}
\label{eq:policyc}
\end{equation}
and our decoding and training objectives change accordingly
(again, we {\em train} the model to catchup using this new policy). 


On the other hand, when  translating from longer source sentences to shorter 
targets, e.g., from English to Chinese,
it is very possible that the decoder finishes generation before
the encoder sees the entire source sentence,
ignoring the ``tail'' on the source side.
Therefore, we need ``reverse'' catchup, i.e., catching up on encoder instead of decoder. 
For example, in English-to-Chinese translation, 
we encode one extra word
every 4 steps, i.e., encoding 5 English words per 4 Chinese words.
In this case, the ``decoding'' catcup frequency $c=r-1=-0.2$ is negative but Eq.~\ref{eq:policyc} still holds.
Note that it works for any arbitrary $c$, such as 0.341, where the catchup pattern is not as easy as ``1 in every 4 steps'', but still maintains a rough frequency
of $c$ catchups per source word.

\begin{figure}[!h]
\centering
\includegraphics[width=7.3cm]{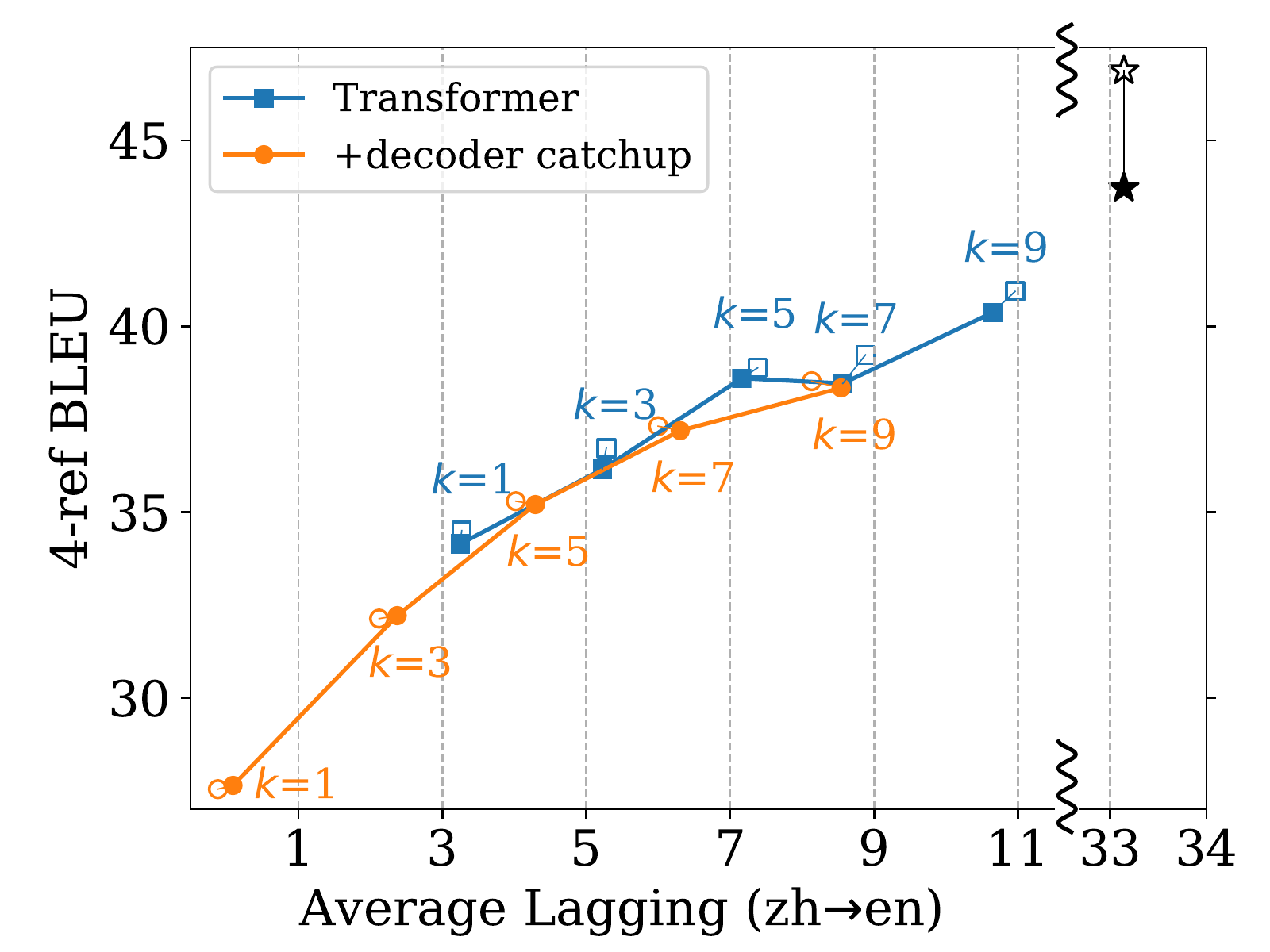} 
\captionof{figure}{
BLEU scores and AL comparisons with different wait-$k$ models on Chinese-to-English on dev set. 
\textcolor{colorC0}{\scriptsize$\square$} and \textcolor{colorC1}{$\circ$} are decoded with tail beam search. {$\bigstar$\ding{73}} and  \textcolor{colorC0}{$\bigstar$\ding{73}} are greedy decoding and beam-search baselines.}
\label{fig:bleuAL_dev1}
\end{figure}

Fig.~\ref{fig:bleuAL_dev1} shows the comparison between wait-$k$ 
model and catchup policy which enables one extra word decoding 
on every $4^{th}$ step. For example, for wait-$3$ policy with 
catchup, the policy is R R (R W R W R W R {\underline{W W}})$^+$ 
W$^+$.

\label{sec:catchup}


\section{Supplemental Material: Evaluations with AP}

\balance
We also evaluate our work using Average Proportion (AP) on both 
de$\leftrightarrow$en and zh$\leftrightarrow$en translation comparing with full sentence translation and \namecite{gu+:2017}.

\begin{figure}[!hb]
\begin{tabular}{cc}
\centering
  \includegraphics[width=7.cm]{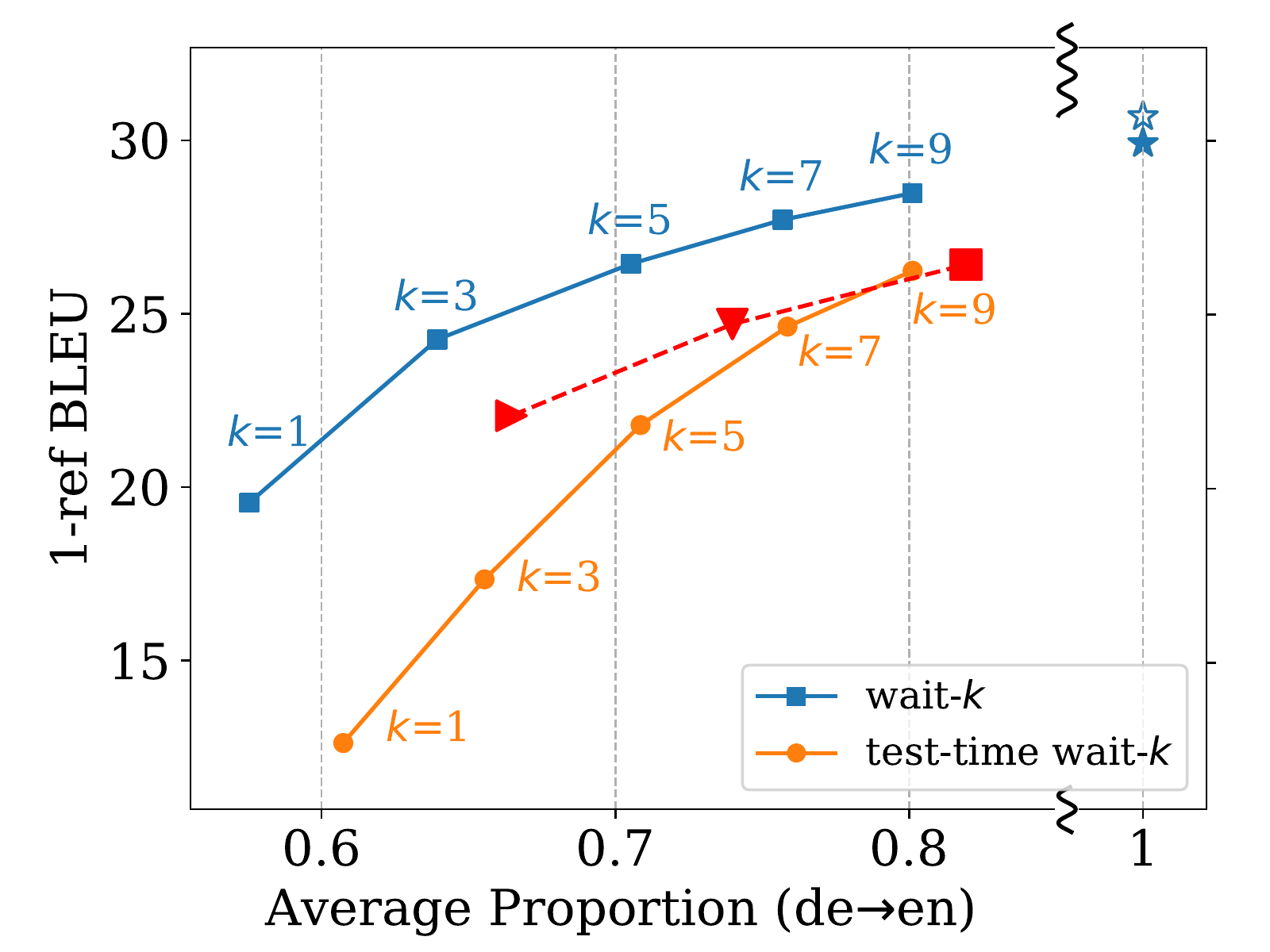} & \\
 \centering
  \includegraphics[width=7.cm]{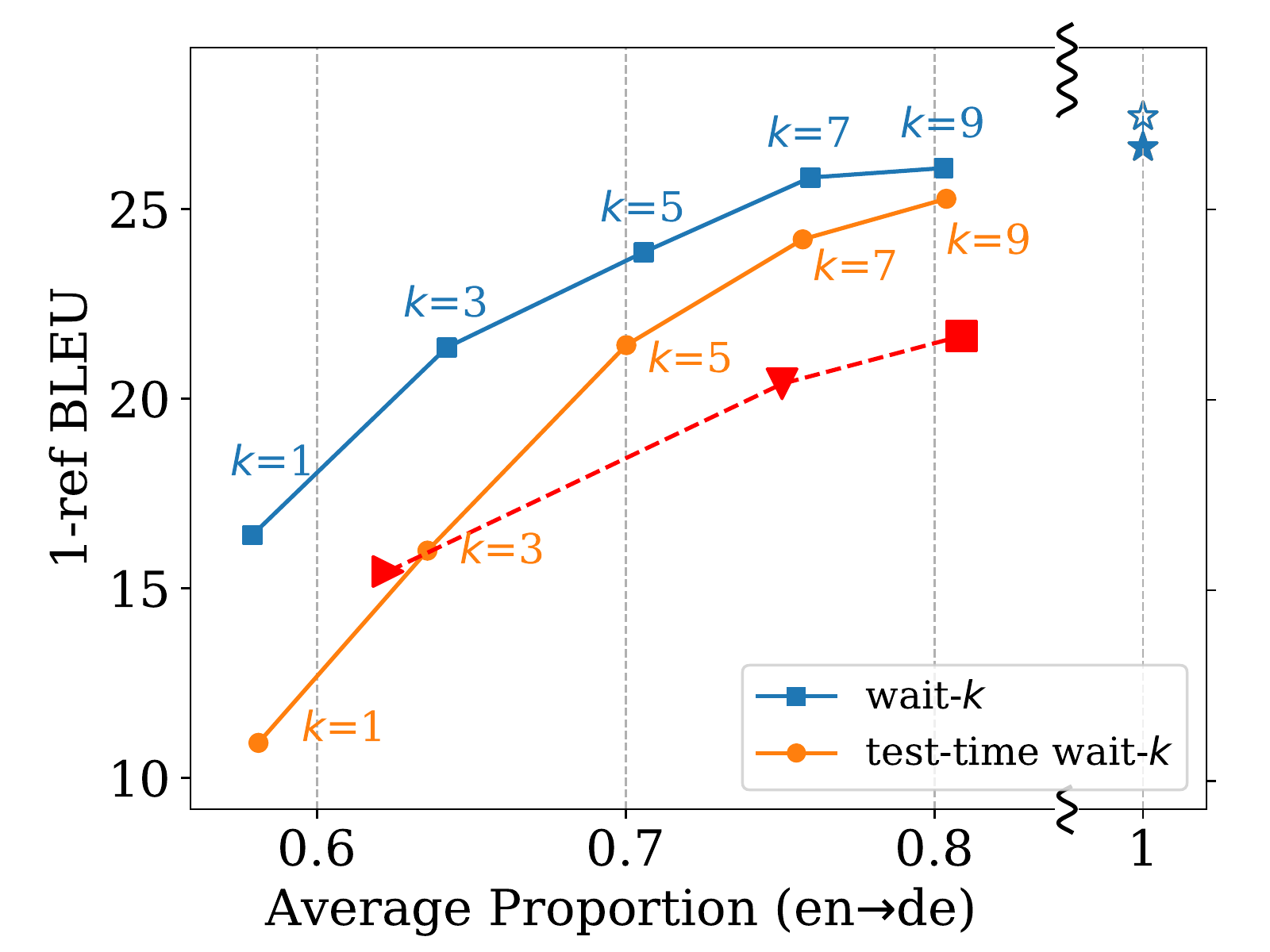}
\end{tabular}\\
\captionof{figure}{Translation quality against AP on de$\leftrightarrow$en simultaneous translation,
  showing wait-$k$ models (for $k$=1, 3, 5, 7, 9), test-time wait-$k$ results, full-sentence baselines, and our reimplementation of \namecite{gu+:2017},
  all based on the same Transformer.
  \textcolor{colorC0}{$\bigstar$\ding{73}}:full-sentence (greedy and beam-search),
  \namecite{gu+:2017}: 
  \textcolor{red}{{$\blacktriangleright$}}:CW=2;
  \textcolor{red}{$\blacktriangledown$}:CW=5;
  \textcolor{red}{$\blacksquare$}:CW=8.
 }
\end{figure}

\begin{figure}[!hb]
\begin{tabular}{cc}
\centering
  \includegraphics[width=7.cm]{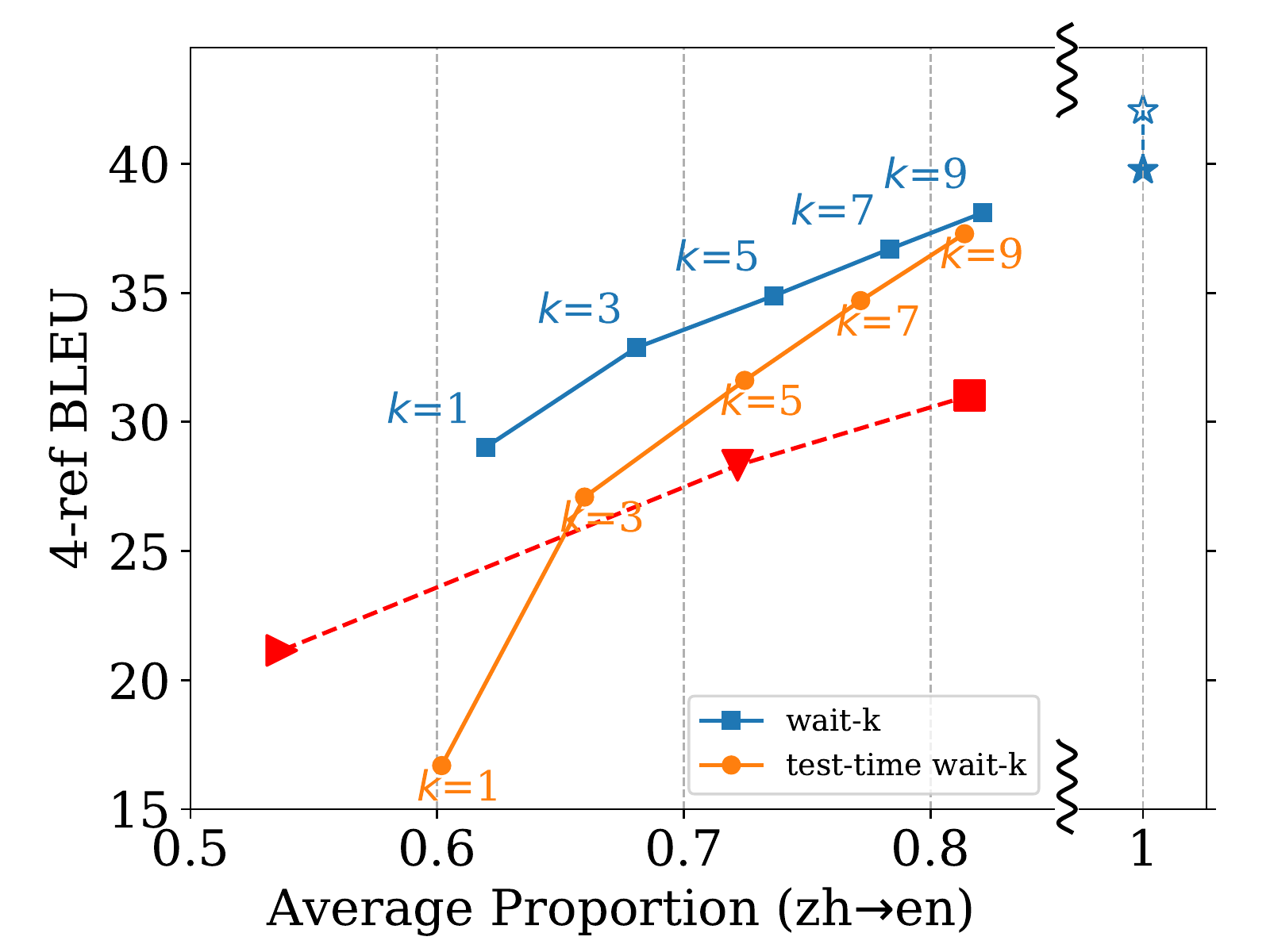} & \\ \centering
  \includegraphics[width=7.cm]{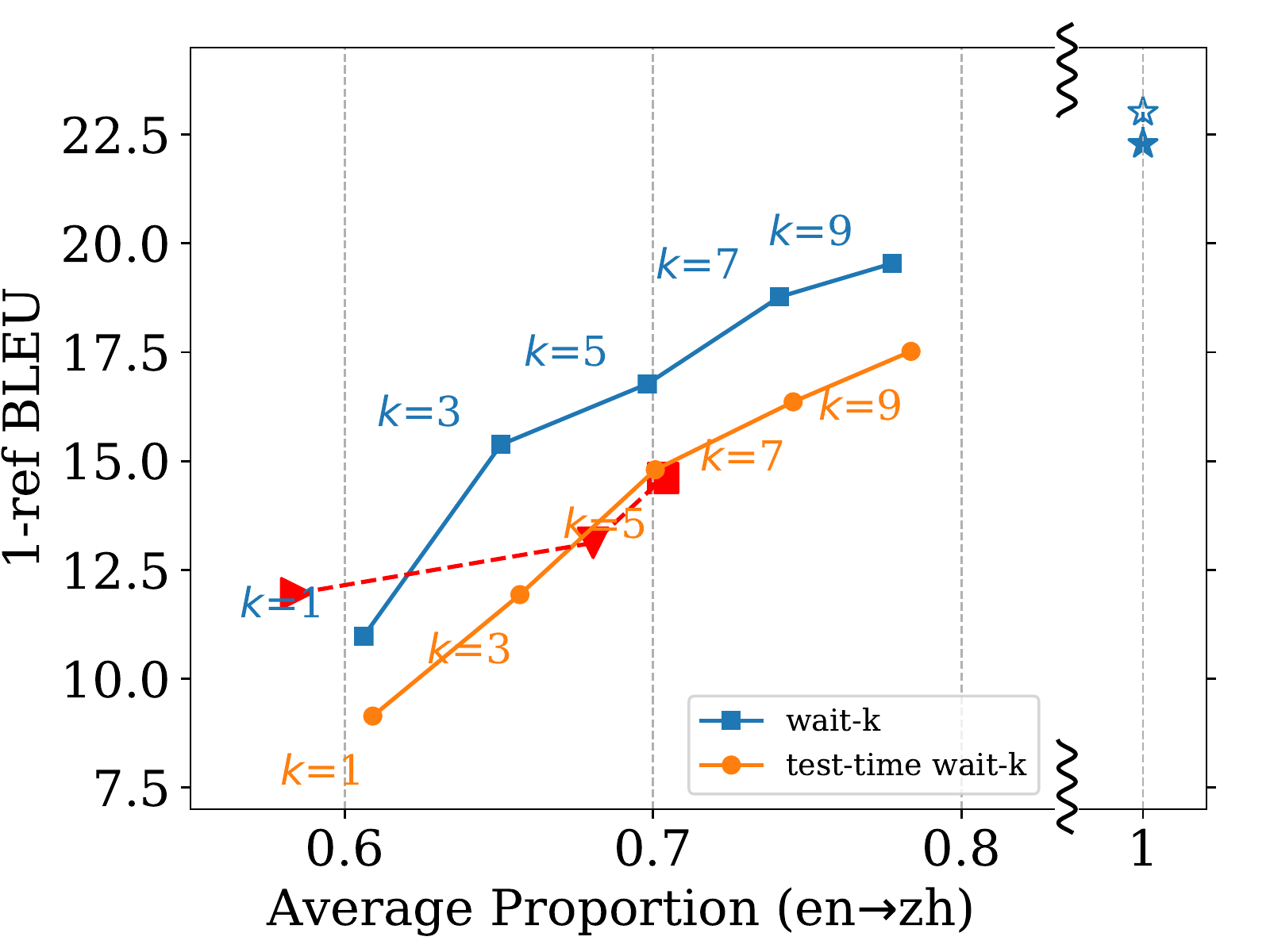}
\end{tabular}\\[-0.3cm]
\captionof{figure}{Translation quality against AP on zh$\leftrightarrow$en simultaneous translation,
  showing wait-$k$ models (for $k$=1, 3, 5, 7, 9), test-time wait-$k$ results, full-sentence baselines, and our reimplementation of \namecite{gu+:2017},
  all based on the same Transformer.
  \textcolor{colorC0}{$\bigstar$\ding{73}}:full-sentence (greedy and beam-search),
  \namecite{gu+:2017}: 
  \textcolor{red}{{$\blacktriangleright$}}:CW=2;
  \textcolor{red}{$\blacktriangledown$}:CW=5;
  \textcolor{red}{$\blacksquare$}:CW=8.
 }
\end{figure}

%
%
%
%
%
\end{CJK}
\end{document}